\documentclass[conference, compsoc]{IEEEtran.1.8b} 

\usepackage{tikz}
\usepackage{amsmath}

\usepackage{filecontents}

\usepackage{multirow}

\usepackage{array}
\usepackage{url}
\usepackage{lscape}
\usepackage{amsmath}
\usepackage{dirtytalk}
\usepackage{color}
\usepackage{subfigure}
\usepackage{listings}
\usepackage{pgf-pie}
\usepackage{booktabs}
\usepackage{hyperref}
\usepackage{xcolor}
\usepackage{framed}
\usepackage{amsfonts}
\usepackage{enumitem}
\usepackage{tikz}
\usepackage{caption}
\usepackage{threeparttable}

\usepackage{algorithm}
\usepackage{algpseudocode}

\graphicspath{{./figures/}}

\newcommand\bsub[1]{\vspace{3pt}\noindent\textbf{#1}}

\definecolor{mygreen}{rgb}{0,0.6,0}
\definecolor{mygray}{rgb}{0.5,0.5,0.5}
\definecolor{mymauve}{rgb}{0.58,0,0.82}

\begin{document}
\title{Low Rank Adaptation for Adversarial Perturbation}
\author{
    \IEEEauthorblockN{
    {Han Liu},
    {Shanghao Shi},
    {Yevgeniy Vorobeychik},
    {Chongjie Zhang},
    {Ning Zhang}
    } 
    Washington University in St. Louis
}






%

\maketitle

\begin{abstract}

Low-Rank Adaptation (LoRA), which leverages the insight that model updates typically reside in a low-dimensional space, has significantly improved the training efficiency of Large Language Models (LLMs) by updating neural network layers using low-rank matrices. Since the generation of adversarial examples is an optimization process analogous to model training, this naturally raises the question: Do adversarial perturbations exhibit a similar low-rank structure?

In this paper, we provide both theoretical analysis and extensive empirical investigation across various attack methods, model architectures, and datasets to show that adversarial perturbations indeed possess an inherently low-rank structure. This insight opens up new opportunities for improving both adversarial attacks and defenses. We mainly focus on leveraging this low-rank property to improve the efficiency and effectiveness of black-box adversarial attacks, which often suffer from excessive query requirements. Our method follows a two-step approach. First, we use a reference model and auxiliary data to guide the projection of gradients into a low-dimensional subspace. Next, we confine the perturbation search in black-box attacks to this low-rank subspace, significantly improving the efficiency and effectiveness of the adversarial attacks.
We evaluated our approach across a range of attack methods, benchmark models, datasets, and threat models. The results demonstrate substantial and consistent improvements in the performance of our low-rank adversarial attacks compared to conventional methods.

\end{abstract}

\thispagestyle{plain}
\pagestyle{plain}


\section{Introduction} \label{sec:intro}

Large language models (LLMs) are advancing rapidly and have become an integral part of modern Internet infrastructure and software applications \cite{vassilev2024adversarial}. Given their growing importance, the security risks associated with LLMs have gained increasing attention, becoming one of the most popular topics in the security research community \cite{carlini2024stealing,liu2024formalizing,zhang2024instruction}. While much of the existing work focuses on the security of LLMs, a less-explored yet intriguing direction is the reverse: how the insights derived from LLM research can inform and benefit security research. One such insight arises from the observation that model updates during LLM fine-tuning often reside in a low-dimensional subspace. This observation led to the development of Low-Rank Adaptation (LoRA), a parameter-efficient fine-tuning technique that introduces low-rank matrices into dense layers to drastically reduce the number of trainable parameters. Rather than updating all weights, LoRA updates only a small number of low-rank parameters, significantly improving training efficiency. Beyond LLMs, LoRA has shown strong applicability across domains such as computer vision \cite{he2023parameter}, multi-task learning \cite{agiza2024mtlora}, and federated learning \cite{cho2023heterogeneous}.

\bsub{Low Rank Opportunity in Adversarial Examples: } 
Since the introduction of adversarial examples \cite{szegedy2013intriguing}, these attacks have gained significant attention in the machine learning (ML) security community \cite{goodfellow2014explaining, madry2017towards, carlini2017towards} as ML systems are increasingly deployed to support critical societal functions. 
Adversarial example generation, like model training, is inherently an optimization process. This raises a natural question: \textit{Do adversarial perturbations also exhibit a low-rank structure?} Answering this question can provide new opportunities for both attack and defense strategies. From the attack perspective, state-of-the-art black-box adversarial attacks, which are among the most practical threat models, often require an excessively large number of queries \cite{cheng2019sign, Vo2022, chen2020hopskipjumpattack}. These high query counts lead to substantial financial costs when targeting commercial APIs and can raise alarms within service providers’ intrusion detection systems \cite{chen2020hopskipjumpattack, suya2020hybrid}. From the defense perspective, adversarial training remains one of the most effective techniques for protecting ML models from adversarial examples \cite{kurakin2022adversarial}. However, such training typically increases computational overhead by an order of magnitude and consumes significantly more memory compared to standard training \cite{shafahi2019adversarial,madry2017towards}.

\bsub{Our Work: }
To answer this question, we revisit adversarial perturbations from a low-rank perspective. Exploring the low-rank property of adversarial perturbations has key differences compared with prior works. Unlike model training, which optimizes model parameters, adversarial example generation focuses on optimizing model inputs. While several studies have examined the role of low-rank properties in adversarial ML \cite{awasthi2020adversarial,nar2019cross,esmaeili2023low,savostianova2024low}, these efforts have primarily explored low-rank properties of natural data and have remained largely empirical without
offering a theoretical analysis of the adversarial perturbations themselves. To address this gap, we first provide a theoretical analysis proving that adversarial perturbations are low-rank. We then conduct a comprehensive empirical rank analysis across a range of attack methods, model architectures, and datasets. Both our theoretical and empirical results consistently reveal that \textit{adversarial perturbations exhibit an inherently low-rank structure.}
This insight opens a new problem space for both attack design and defense development. In particular, by constraining the search for adversarial perturbations to a low-rank subspace, we can significantly improve the efficiency and effectiveness of adversarial example generation. Furthermore, integrating low-rank optimization into adversarial training can reduce significant memory overhead. While we provide a theoretical foundation for these benefits, our primary focus is to demonstrate how leveraging low-rank subspaces can improve the efficiency and efficacy of black-box adversarial attacks. However, this approach introduces two key challenges:

\vspace{2pt}\noindent\textit{C1. Constructing the Low-Rank Subspace in Black-Box Setting:} In black-box setting, the adversary lacks direct access to the gradients needed to update adversarial examples, making it impossible to directly decompose these updates into a low-rank matrix, as is done in LoRA \cite{hulora2022}. To address this, we leverage the insight that different models share similarities and correlations in their low-rank subspaces. We assume the adversary has access to auxiliary data, which does not share the same distribution or labels as the training data of the target model, as well as a reference model with a different architecture and training distribution. Using this auxiliary dataset and reference model, we generate gradient vectors. To mitigate the noise in these gradients, we apply explainable machine learning techniques to identify their critical components. These components are then used to train an autoencoder, which constructs a low-rank subspace from the learned latent space.

\vspace{2pt}\noindent\textit{C2. Utilizing the Low-Rank Subspace for Adversarial Attacks:} The second challenge is effectively utilizing the constructed low-rank subspace to execute adversarial attacks. Given the sophistication of existing black-box attacks, our goal is to integrate the low-rank subspace into these attacks to improve their effectiveness and efficiency. The difficulty arises from the diversity of mechanisms employed by different attacks, making it challenging to develop a generalizable framework that works across various strategies. To overcome this, we first abstract the core components of different black-box attacks and identify key integration points for incorporating the low-rank subspace. A common characteristic among these attacks is their reliance on random vectors for perturbation direction estimation. By restricting these random vectors to the low-rank subspace, we constrain the adversarial optimization process within this subspace, enhancing the attack efficiency.

\bsub{Experiment and Results: }
We conduct a comprehensive evaluation of the effectiveness of our low-rank attack framework by applying it to a range of advanced attacks, including score-based and decision-based black-box attacks, across both untargeted and targeted settings. These experiments span various benchmark datasets, model architectures, and robust models. Our results show a significant improvement in both the effectiveness and efficiency of the low-rank attacks compared to the original attacks. Additionally, we systematically examine the impact of auxiliary datasets and reference models, demonstrating that our framework remains effective even when these resources differ substantially from the victim model and its training data.




\bsub{Contributions:} Our contributions are outlined as follows:
\begin{itemize} [leftmargin=10pt, itemsep=0mm, parsep=0mm]
    \item We conduct a systematic investigation, both theoretical and empirical, into the rank properties of adversarial perturbations. Our analysis reveals a fundamental insight that adversarial perturbations inherently exhibit a low-rank structure.

    \item We propose a novel low-rank attack framework that leverages this low-rank property for crafting adversarial examples. Our framework is designed to be both highly effective and efficient, enhancing the performance of existing adversarial attacks. 

    \item Through extensive experiments, we demonstrate significant improvements in both the efficacy and efficiency of our low-rank attacks compared to baselines. These improvements are observed across various attack techniques, datasets, model architectures, and threat models. 
\end{itemize}

\section{Background} \label{sec:background}

\subsection{Low Rank Model Training}
\bsub{Low-Rank Adaptation. }
Low-Rank Adaptation (LoRA) \cite{hulora2022} has garnered significant research attention and widespread application since its introduction by Hu et al. This approach builds upon the observation that updates made during fine-tuning of LLMs exhibit a low intrinsic rank. LoRA proposes to incrementally update the pre-trained weights by using the product of two low-rank matrices. Specifically, for a linear layer $W \in \mathbb{R}^{m \times n}$, LoRA models the weight update $\Delta W \in \mathbb{R}^{m \times n}$ using a low-rank decomposition, expressed as $BA$, where $B \in \mathbb{R}^{m \times r}$ and $A \in \mathbb{R}^{r \times n}$ are two low-rank matrices, with $r \ll \min(m, n)$. Then the fine-tuned weight $W'$ can be represented as:

\begin{equation} \label{eq1:low_rank}
W' = W_0 + \Delta W = W_0 + BA
\end{equation}

During the fine-tuning process, the original weight $W_0$ is frozen, and only the low-rank matrices $A$ and $B$ are updated. Additionally, the matrix $A$ is typically initialized using the uniform Kaiming distribution \cite{he2015delving}, while $B$ is initially set to zero, resulting in $\Delta W = BA$ being zero at the start of training.
LoRA offers significant benefits for model training. First, it substantially reduces GPU memory utilization. During the training process, optimizers are memory-intensive. For example, in the case of the Adam optimizer \cite{kingma2014adam}, a weight matrix $W_0 \in \mathbb{R}^{m \times n}$ requires storage for $W_0$, the first moment estimate $M \in \mathbb{R}^{m \times n}$, and the second moment estimate $N \in \mathbb{R}^{m \times n}$, resulting in a total memory usage of approximately $3mn$, nearly three times the original parameter memory. This issue is particularly problematic for large-scale models, such as LLMs, which have extensive parameter counts. However, since LoRA freezes the original weight $W_0$ and only updates $A$ and $B$, the total memory usage is reduced to $3(m+n)r$. Given that $r \ll \min(m, n)$, this leads to a significant decrease in memory utilization. Furthermore, as the frozen parameters do not require gradient calculations during backpropagation, computational costs are also reduced, further boosting training efficiency. LoRA has demonstrated success across various domains and tasks, including LLMs \cite{dettmers2024qlora}, computer vision \cite{he2023parameter}, and multi-task learning \cite{agiza2024mtlora}, etc.


\bsub{Singular Value Decomposition. } Singular Value Decomposition (SVD) is one of the most common methods to analyze the rank of a matrix. It decomposes a given matrix \( A \in \mathbb{R}^{m \times n} \) into three matrices \( U \in \mathbb{R}^{m \times m} \), \( \Sigma \in \mathbb{R}^{m \times n} \), and \( V^T \in \mathbb{R}^{n \times n}\):

\begin{equation}
    A = U \Sigma V^T 
\end{equation}
where \( U \) and \( V \) are orthogonal matrices containing the left and right singular vectors, and \( \Sigma \) is a diagonal matrix containing the singular values of \( A \), which indicate the rank and intrinsic dimensionality of the matrix. SVD can also be used for low-rank approximation by truncating the smaller singular values in \( \Sigma \), allowing for an efficient representation of \( A \) with reduced dimensionality.

\subsection{Adversarial Attacks} \label{subsec:adv-attack}


Adversarial attacks can be classified both by their objectives and by the threat model they assume. From the objective perspective, these attacks are divided into untargeted and targeted. Untargeted attacks seek to make a model misclassify an input into any incorrect label, while targeted attacks aim to misclassify an input into a specific label chosen by the adversary. From the threat model perspective, adversarial attacks are commonly split into white-box and black-box attacks. White-box attacks assume that the adversary has complete access to the target model's internal information, such as its architecture, parameters, and gradients. However, such comprehensive access is typically not feasible in practical scenarios, which has shifted the focus of latter research towards black-box attacks, where the adversary lacks direct access to the model's internal information but can observe the model's inputs and corresponding outputs. Black-box attacks can be further subdivided into score-based and decision-based attacks. In score-based attacks, the adversary has access to the confidence scores, while in decision-based attacks, the adversary only knows the predicted labels.
nevunlg

Earlier studies mainly focus on white-box attacks, such as FGSM \cite{goodfellow2014explaining}, BIM \cite{kurakin2018adversarial}, PGD \cite{madry2017towards}, and the C\&W \cite{carlini2017towards}. 
Further research expanded these attacks to more practical threat models, assuming the adversary has access to the full or partial confidence scores returned by the model \cite{chen2017zoo,uesato2018adversarial,andriushchenko2020square,ilyas2018prior}. 
To enhance attack stealthiness, a significant line of research has formulated adversarial example generation as a constrained optimization problem. The goal is to craft an adversarial example $x_{adv}$ such that the perturbation from the original example $x$ is bounded by \(\| x_{adv} - x \|_p \leq \epsilon\), where \( \epsilon \) is the perturbation budget. For instance, Ilyas et al. \cite{ilyas2018prior} propose a bandit optimization framework that effectively integrates gradient priors into the constrained optimization process for generating black-box adversarial examples. 
Another major line of research focuses on improving attack stealthiness by identifying the smallest possible perturbation maintaining adversarial effectiveness, as seen most of decision-based attacks.
%
Decision-based attacks, which could only access the highest confidence class label, are considered to have the most practical threat models since pervasive real-world applications typically provide no information other than the label \cite{chen2020hopskipjumpattack,fu2022autoda,cheng2019sign,cheng2018query}. Due to the minimal information available, these attacks usually initialize with a reference image that satisfies the adversary's goal and then focus on minimizing the size of the perturbation to the original image. 
Opt \cite{cheng2018query} formulates the decision-based attack as an optimization problem that finds a direction that could produce the shortest distance to the decision boundary. Sign-Opt \cite{cheng2019sign} improves upon Opt by using the sign($\cdot$) operation to enable a more query-efficient gradient estimation. HSJA \cite{chen2020hopskipjumpattack} moves the candidate sample along the gradient direction and projects it back to the decision boundary iteratively to find adversarial examples with minimum perturbation. Additionally, RamBoAttack \cite{Vo2022} proposes Block Descent to avoid entrapment in local minima and misdirection from noisy gradients during the gradient estimation process.

\section{Low Rank Adversarial Perturbation} \label{sec:low_rank}

\begin{figure*}[h]
  \centering
  
  \begin{minipage}{0.195\textwidth}
   \includegraphics[width=\linewidth]{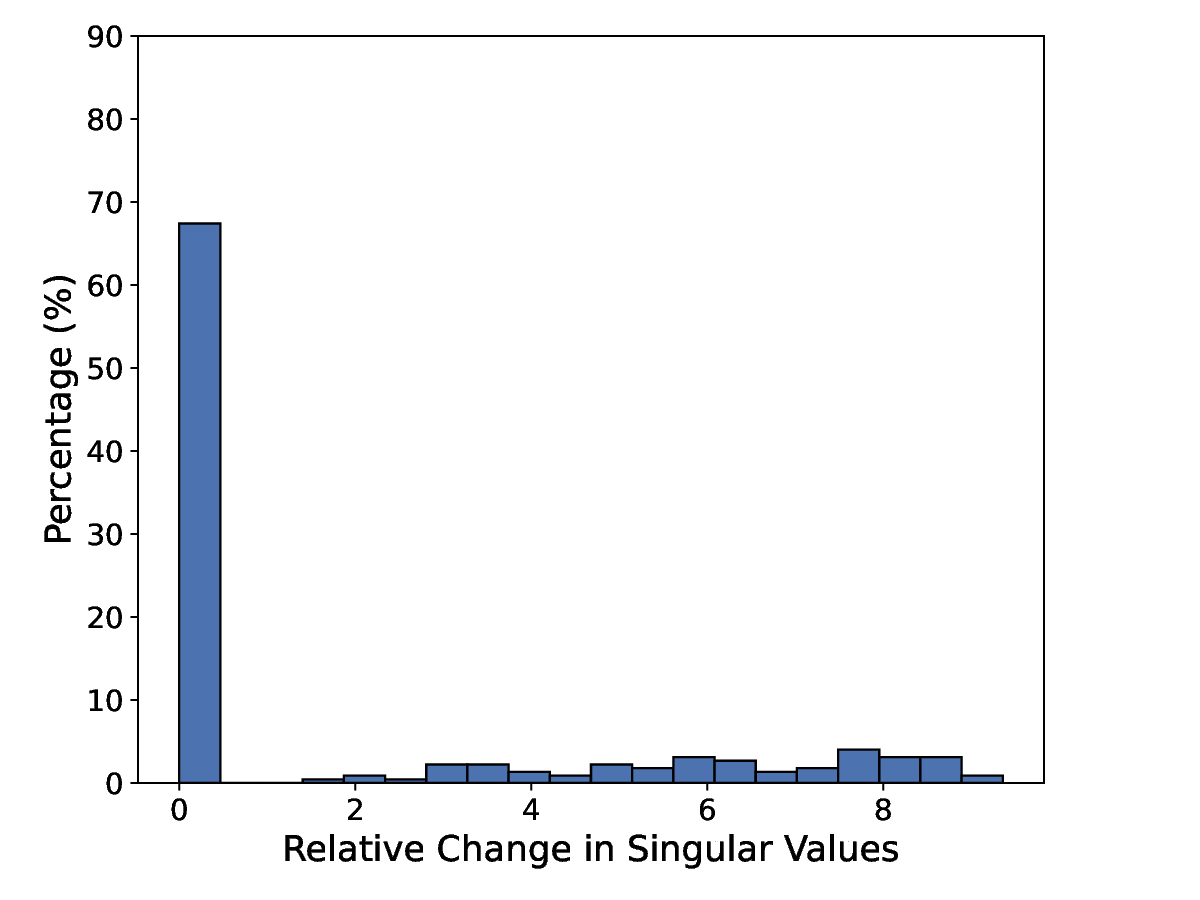}\\
    \includegraphics[width=\linewidth]{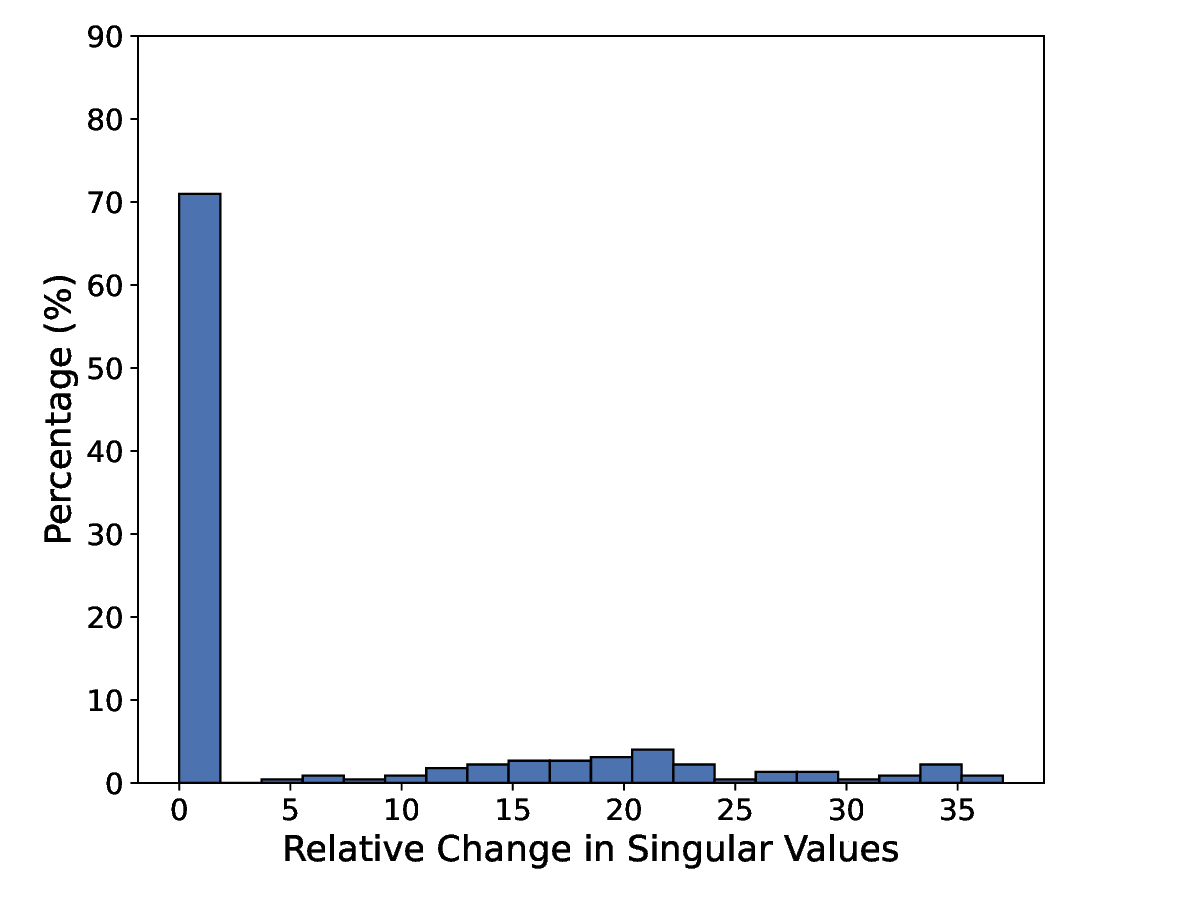}
    \caption*{(a) ImageNet/ResNet}
  \end{minipage}
  \hfill
   \begin{minipage}{0.195\textwidth}
    \includegraphics[width=\linewidth]{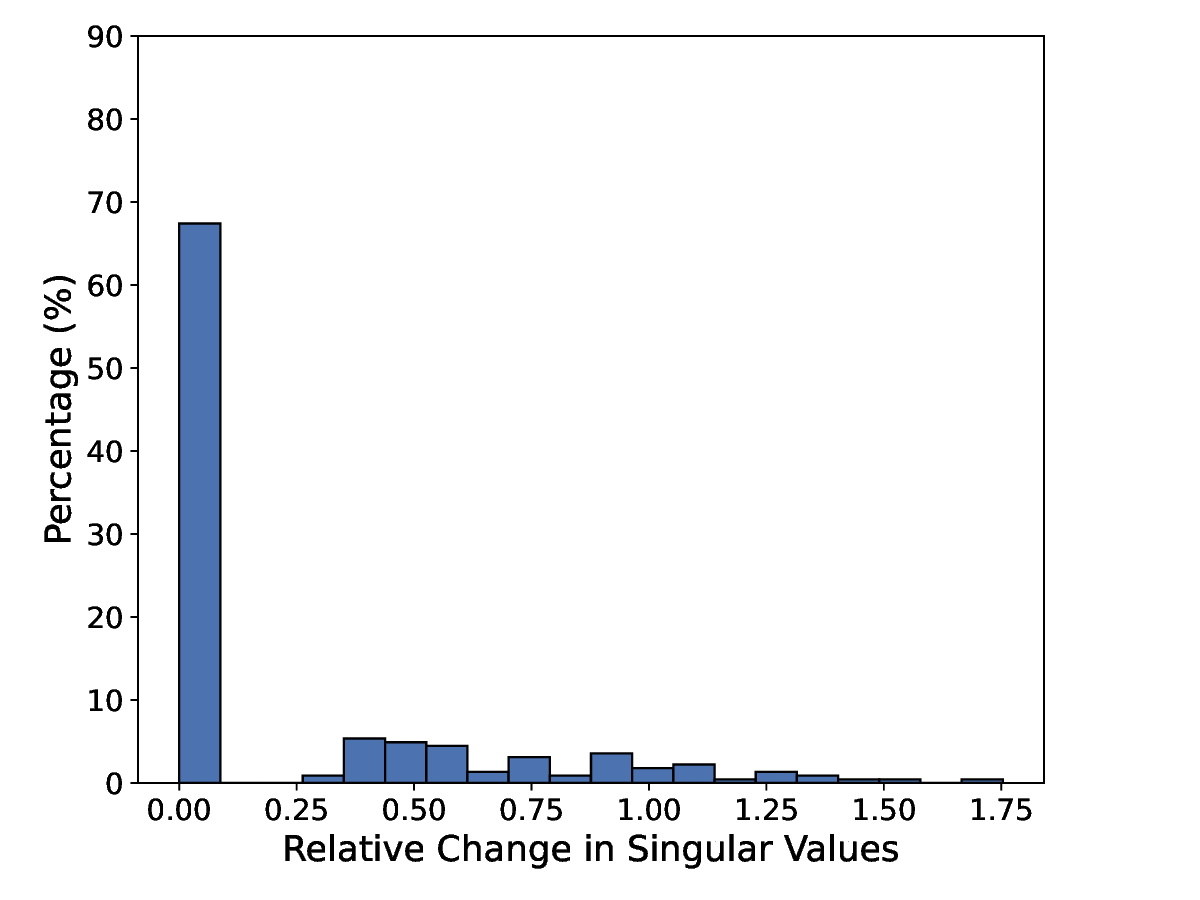}\\
    \includegraphics[width=\linewidth]{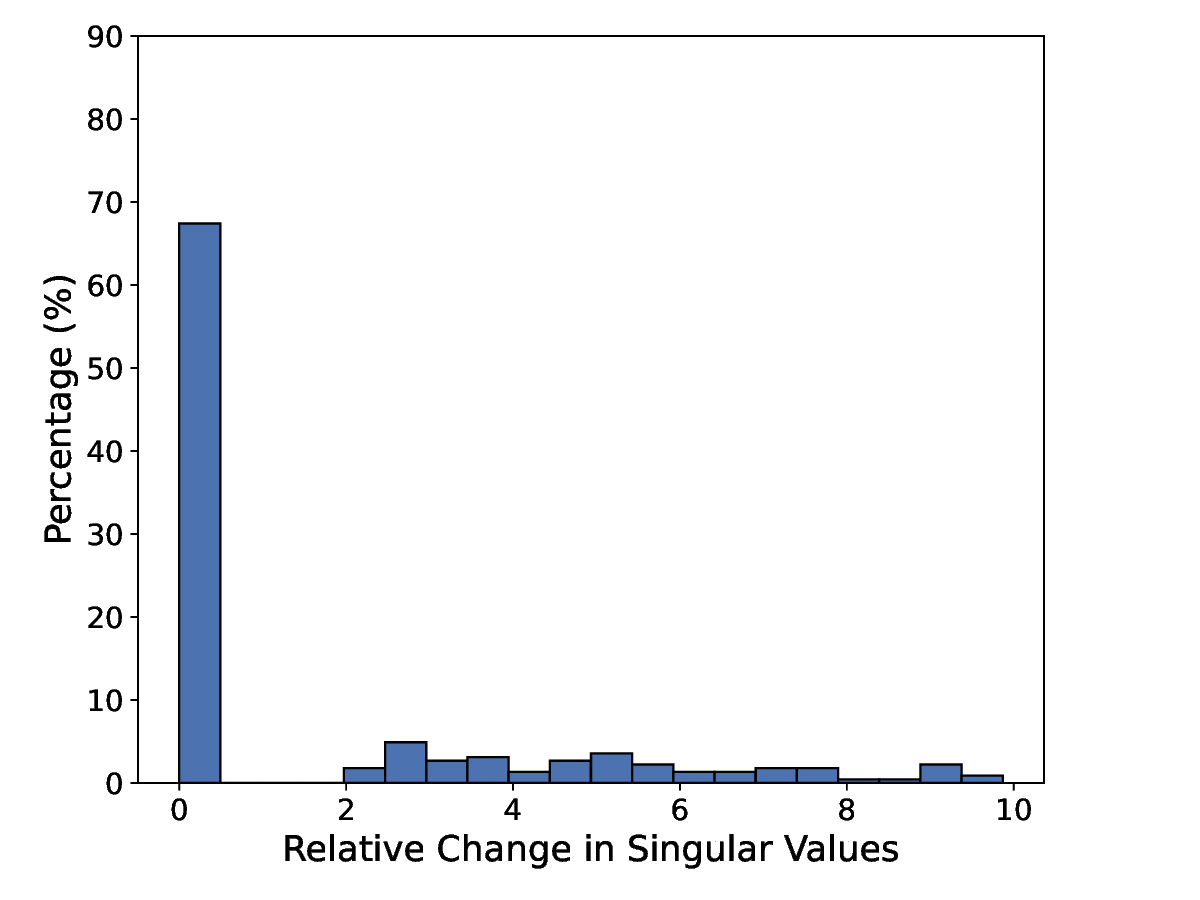}
    \caption*{(b) ImageNet/EffNet}
  \end{minipage}
  \hfill
   \begin{minipage}{0.195\textwidth}
    \includegraphics[width=\linewidth]{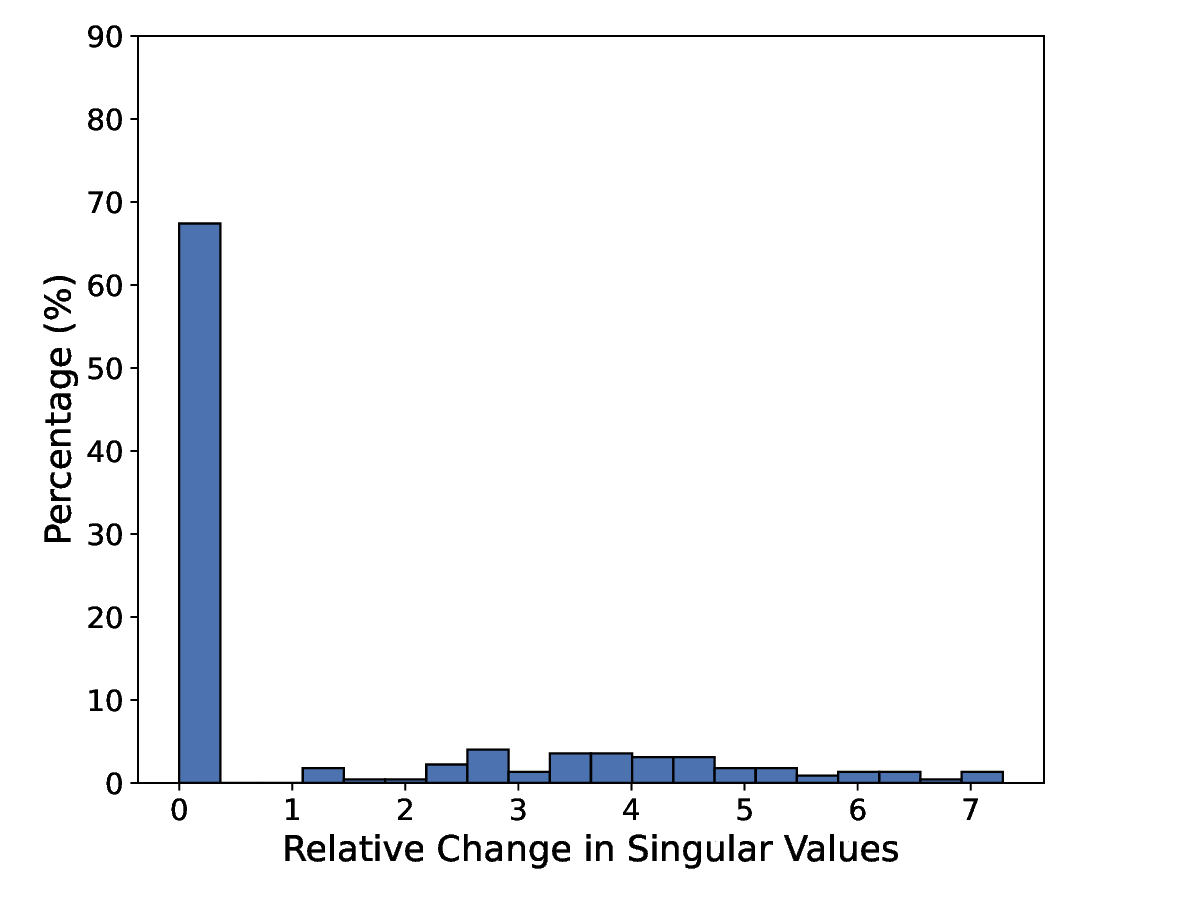}\\
    \includegraphics[width=\linewidth]{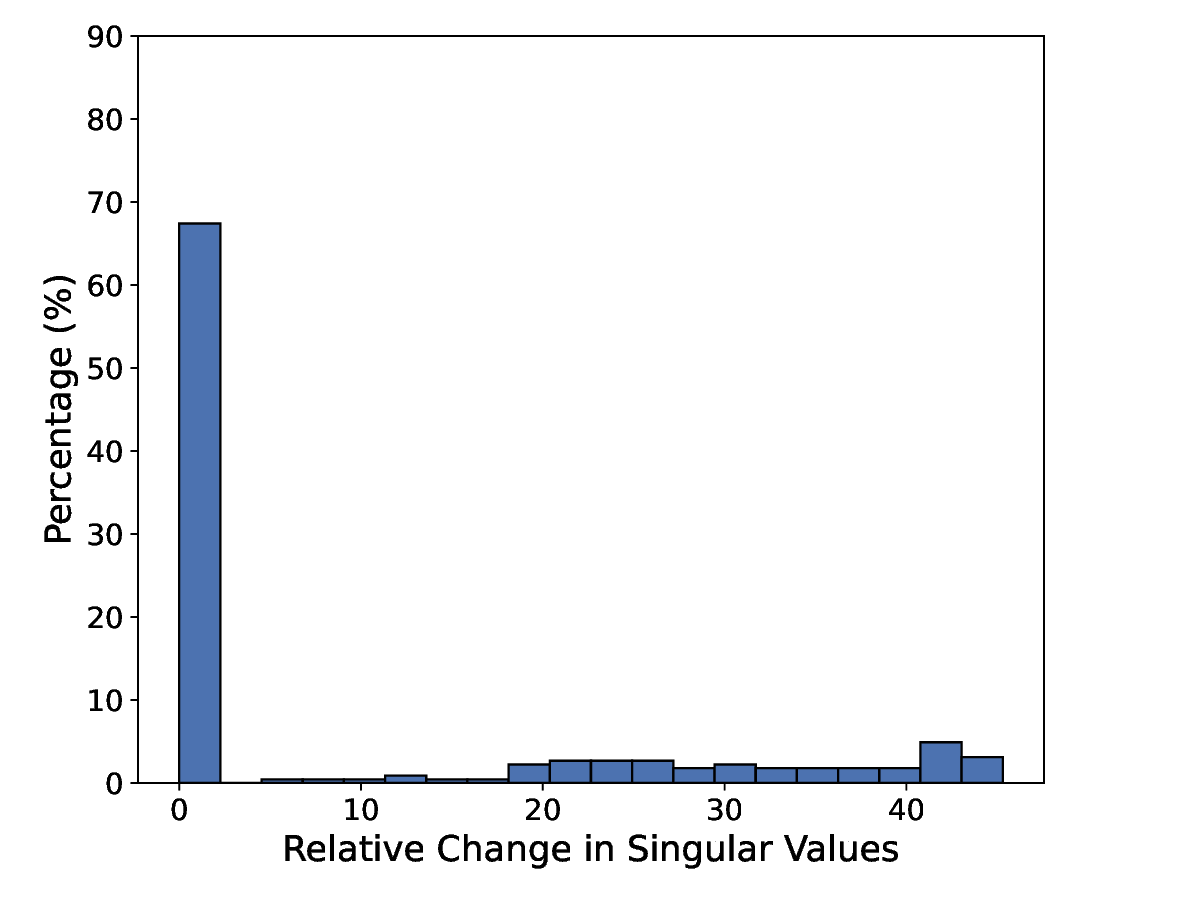}
    \caption*{(c) ImageNet/ViT}
  \end{minipage}
  \hfill
    \begin{minipage}{0.195\textwidth}
    \includegraphics[width=\linewidth]{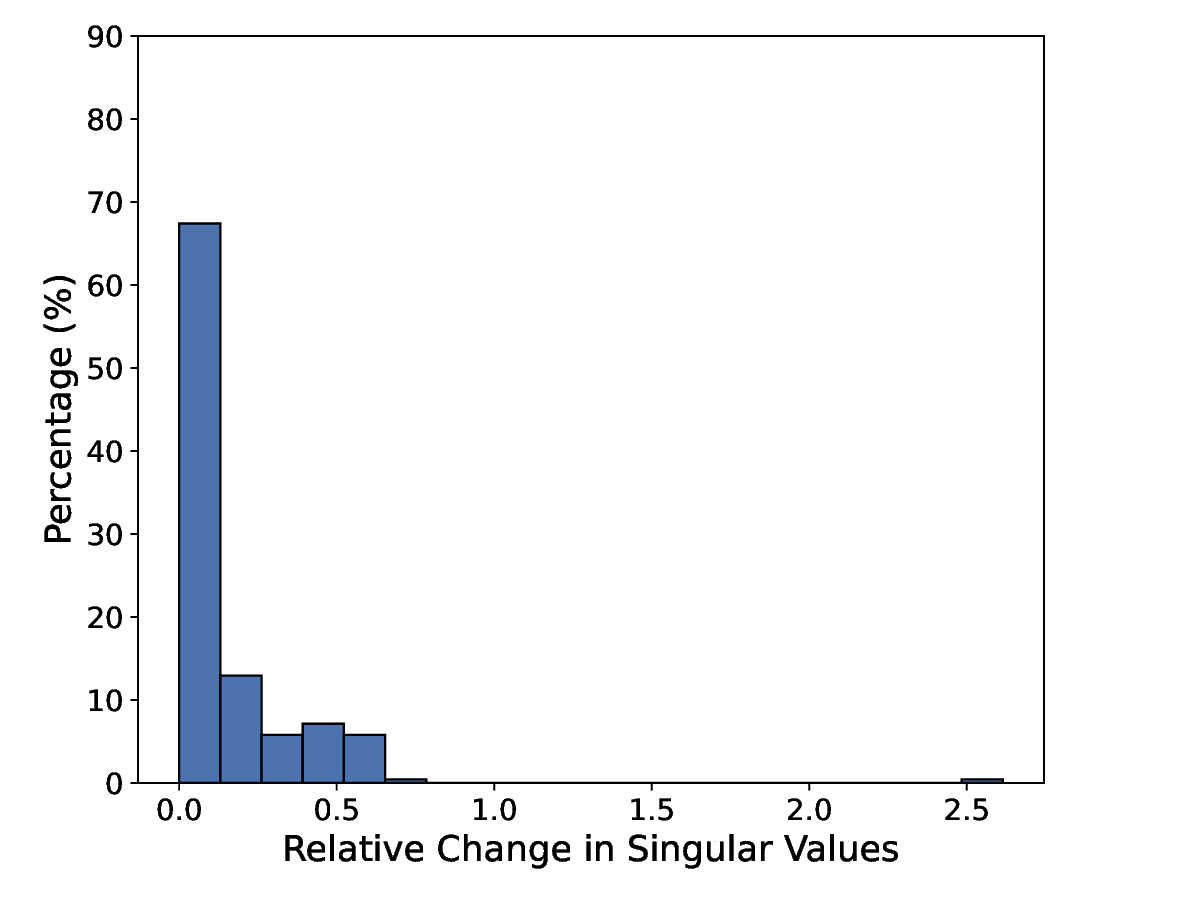}\\
    \includegraphics[width=\linewidth]{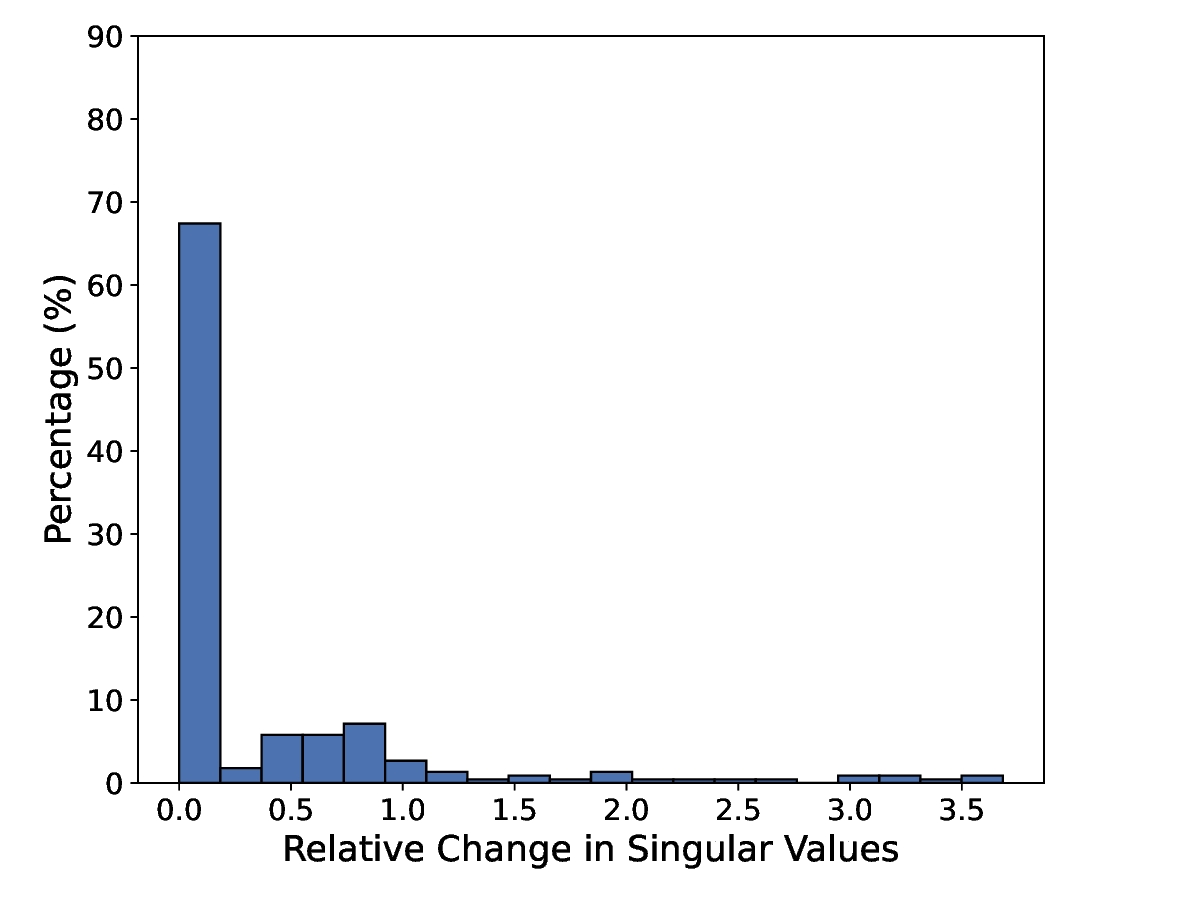}
    \caption*{(d) ImageNet/Robust }
  \end{minipage}
    \hfill
    \begin{minipage}{0.195\textwidth}
    \includegraphics[width=\linewidth]{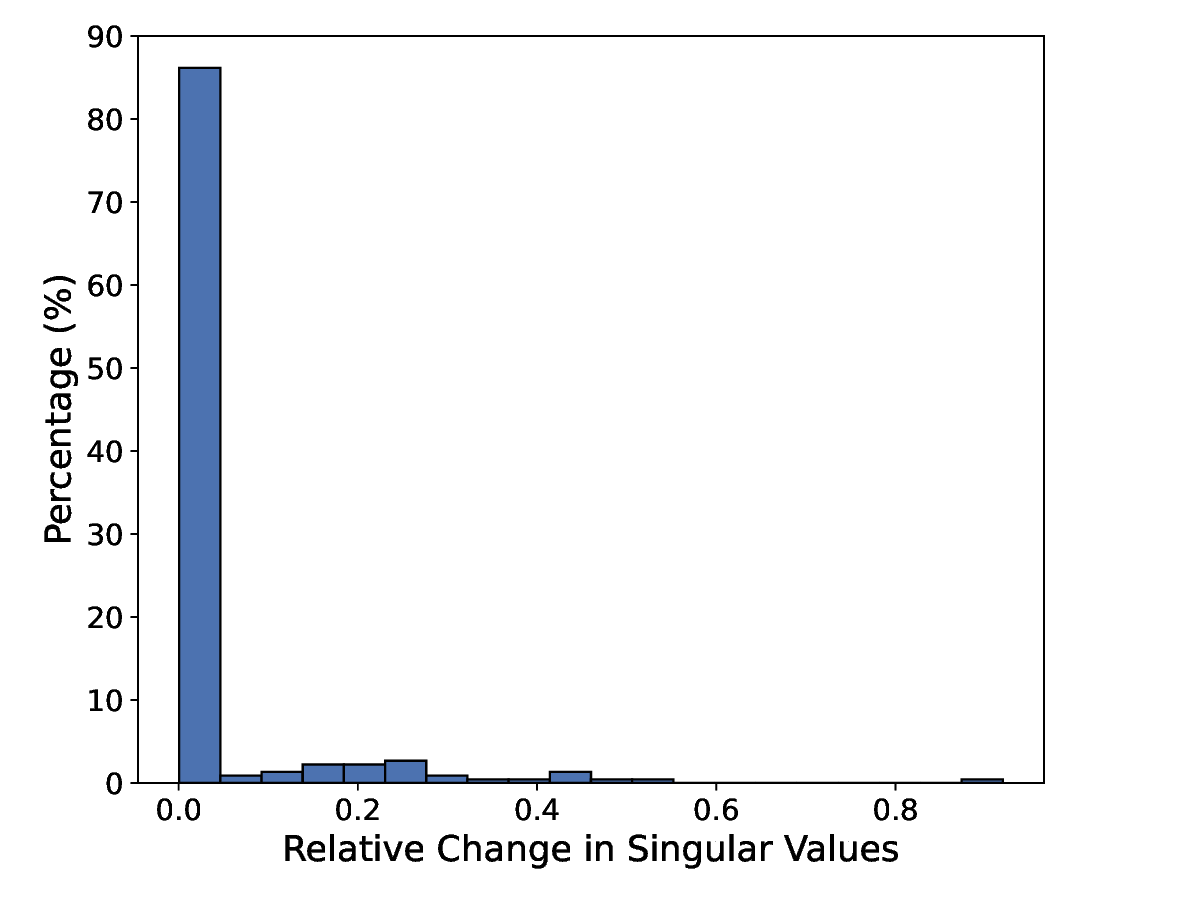}\\
    \includegraphics[width=\linewidth]{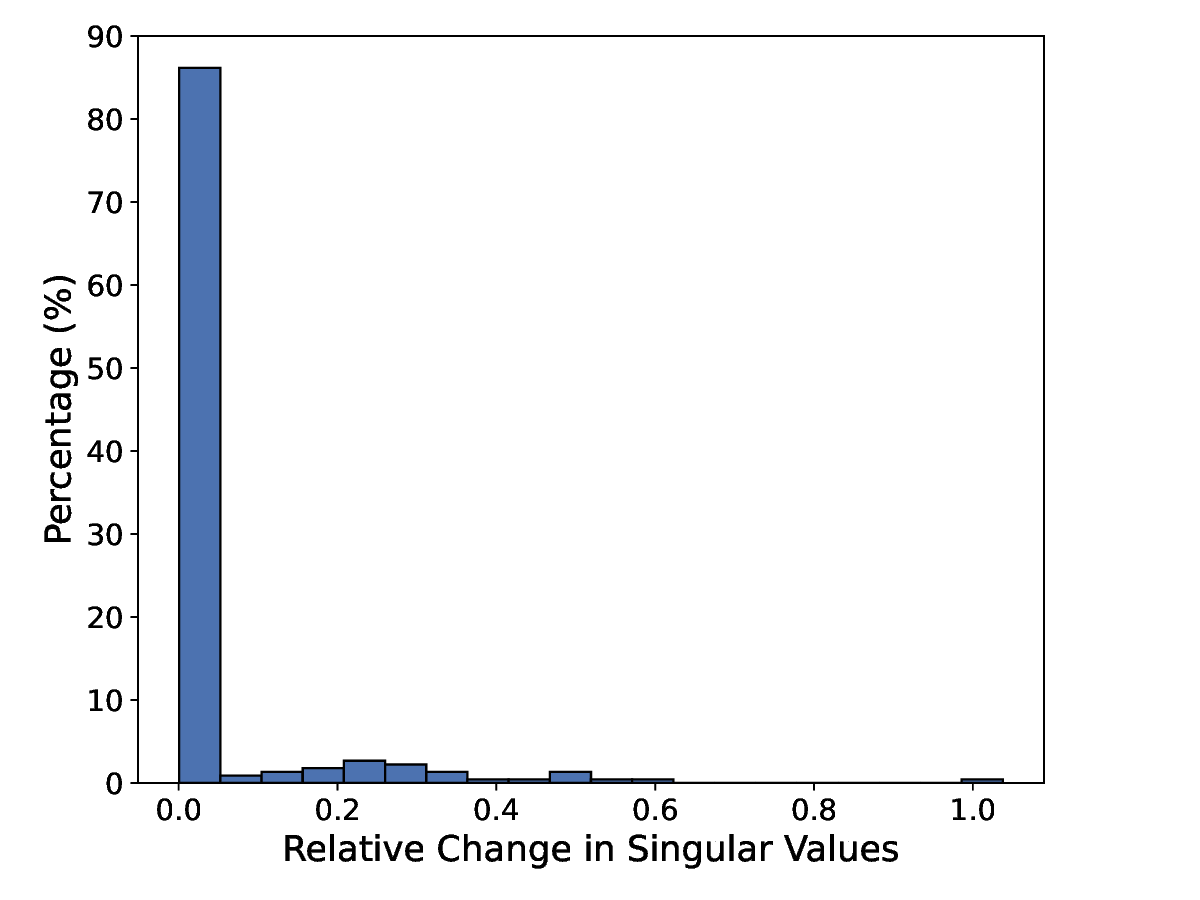}
    \caption*{(e) CUB-200/ResNet}
  \end{minipage}
  \caption{The relative magnitude change of singular values across different adversarial attacks, model architectures and datasets. The first row corresponds to the PGD attack, while the second row represents the C\&W attack. The first three columns display results for models trained on the ImageNet dataset with ResNet-50, EfficientNet, and Vision Transformer, respectively. The fourth column shows results for adversarially trained WideResNet-50 models on ImageNet using the method proposed by Salman et al. The fifth column presents results for ResNet-50 trained on the CUB-200 dataset.}
  \label{fig:low_rank_analysis}
\end{figure*}

\bsub{Motivation. }
The observation and utilization of low-rank properties have notably enhanced the efficiency of optimizing LLMs. Given that generating adversarial examples is inherently an optimization process, it naturally leads to a question: \textit{do adversarial perturbations also exhibit a low-rank property?}

This investigation differs from previous research in two key aspects. Firstly, unlike model training, which optimizes model parameters, generating adversarial perturbations involves optimizing the model's inputs. Secondly, while existing studies on adversarial ML have investigated low-rank structures \cite{awasthi2020adversarial,nar2019cross,esmaeili2023low,entezari2020all,savostianova2024low}, their primary focus has been on leveraging the low-rank nature of natural data, and these efforts have remained largely empirical without offering theoretical analysis of the adversarial perturbations themselves.
For instance, Nar et al. \cite{nar2019cross} showed that low-rank features in training data, when combined with cross-entropy loss, result in smaller margins between a classifier's decision boundary and training points, leading to differential training schemes aimed at improving adversarial robustness. Similarly, Awasthi et al. \cite{awasthi2020adversarial} leveraged low-rank data representations to strengthen certified robustness. Esmaeili et al. \cite{esmaeili2023low} exploited the observation that most images can be decomposed into a low-rank background and a sparse foreground, devising an efficient untargeted adversarial attack by restricting perturbations to this sparse component, thereby improving query efficiency.
However, there remains a notable gap: \textit{existing research lacks systematic theoretical and empirical analysis regarding whether adversarial perturbations themselves inherently exhibit low-rank structures}. Addressing this gap can unlock significant new insights and facilitate improvements in the efficiency and effectiveness of both adversarial attacks and defenses, as will be discussed in detail later in this section.

In the remainder of this section, we first provide a theoretical proof for the low-rank nature of adversarial perturbations. We then conduct a comprehensive empirical rank analysis of perturbations across diverse attack methods, model architectures, and datasets. Finally, we discuss the broader implications of this property from both the attack and defense perspectives.

\bsub{Theoretical Analysis. }
We now theoretically prove the low-rank nature of adversarial perturbations. Consider a neural network classifier \( f \) mapping an input vector \( x \in \mathbb{R}^d \) (e.g., image pixels) to logits corresponding to \( C \) classes, \( f(x) \in \mathbb{R}^{C} \). The Jacobian of \( f \) w.r.t. the input \( x \) is defined as:
\begin{equation}
    J(x) = \frac{\partial f}{\partial x} \in \mathbb{R}^{C \times d}.
\end{equation}

For simplicity, we analyze the scenario of an untargeted attack, which aims to solve the following optimization goal:
\begin{equation} \label{eq:adv_obj}
\min_{\delta} \|\delta\|_2 \quad \text{s.t.} \quad f_y(x + \delta) \leq \max_{k \neq y} f_k(x + \delta).
\end{equation}

We denote $\hat{k}$ as the class closest to surpassing the correct class $y$ at the original input $x$, i.e.,
\begin{equation}
\hat{k} = \arg\max_{k \neq y} f_k(x).
\end{equation}

Under the general assumption that neural networks exhibit piecewise affine behavior~\cite{montufar2014number}, we employ a first-order Taylor approximation for a small perturbation $\delta$:
\begin{equation}
f(x + \delta) \approx f(x) + J(x)\delta.
\end{equation}

Substituting this approximation into Eq. \ref{eq:adv_obj} yields:
\begin{equation}
f_{\hat{k}}(x) + J_{\hat{k}}(x)\delta \geq f_y(x) + J_y(x)\delta.
\end{equation}

Define $\boldsymbol{w} = J_{\hat{k}}(x) - J_y(x)$ and $m = f_y(x) - f_{\hat{k}}(x)$, leading to the linear constraint:
\begin{equation}
\boldsymbol{w}^\top \delta \geq m.
\end{equation}

Solving this constrained optimization problem using the method of Lagrange multipliers~\cite{bertsekas2014constrained} yields the unique minimal-norm solution:
\begin{equation}
\delta^\star = \frac{m}{\|\boldsymbol{w}\|_2^2} \boldsymbol{w}.
\end{equation}

Since $\boldsymbol{w}$ is itself a linear combination of two rows of $\boldsymbol{J}$, we have $\text{span}\{\boldsymbol{w}\} \subseteq \text{row}(\boldsymbol{J}(x))$, meaning $\delta^\star$ resides within $\text{row}(\boldsymbol{J}(x))$. This subspace has dimension at most $C$. Given that the dimensionality of the input $d$ is typically much larger than the number of output classes $C$, it follows that $\text{rank}(\boldsymbol{J}(x)) \leq C \ll d$. Thus, the solution $\delta^\star$ is inherently low-rank.
Furthermore, for gradient-based iterative attacks (e.g., PGD, C\&W attacks), the gradient update at each step is computed as:
\begin{equation} \label{eq:iterative_defense}
\nabla_x \mathcal{L}(f(x + \delta_t), y) = \boldsymbol{J}(x + \delta_t)^\top \boldsymbol{v}_t,
\end{equation}
where $\boldsymbol{v}_t = \left. \nabla_z \mathcal{L}(z) \right|_{z = f(x)} \in \mathbb{R}^C$, and $\mathcal{L}$ denotes the loss function. Thus, each gradient update is expressed as a linear combination of at most $C$ columns of $\boldsymbol{J}^\top$. Consequently, every step of optimization remains confined to a subspace of dimensionality at most $C$ within $\mathbb{R}^d$. This observation reveals new opportunities for advancements in adversarial ML, such as more efficient adversarial training, which we will discuss later in this section.

\bsub{Rank Analysis. }
To empirically analyze the rank of updates during adversarial example generation, we employ Singular Value Decomposition (SVD), a widely used method for rank analysis. Our systematic study encompasses representative adversarial attacks (PGD \cite{madry2017towards} and C\&W \cite{carlini2017towards} Attacks), various model architectures (ResNet-50 \cite{he2016deep}, EfficientNet \cite{tan2019efficientnet}, and Vision Transformer (ViT) \cite{dosovitskiy2020image}, which is the current state-of-the-art architecture for image classification), different model training methods (normal training and adversarial training using the representative method proposed by Salman et al. \cite{salman2020adversarially}), and multiple datasets (ImageNet \cite{deng2009imagenet} and CUB-200 \cite{wah2011caltech}). For each scenario, we calculate the average relative magnitude change of singular values across channels for 1000 randomly selected images. To visualize the distribution of these changes in adversarial example updates, we plot the histograms.

The results are given in Figure \ref{fig:low_rank_analysis}. We observe that approximately 70\% of the relative singular value changes are concentrated around zero for ImageNet, while about 80\% for CUB-200. This indicates that adversarial perturbations predominantly affect a small subset of singular values rather than spreading uniformly throughout the high-dimensional input space. In other words, adversarial perturbations largely occupy a few dominant directions and are thus intrinsically low-rank.
Additionally, two noteworthy observations emerge from our study. First, the singular value changes in adversarially trained models (Figure \ref{fig:low_rank_analysis} (d)) are more tightly clustered around zero compared to those in normally trained models (Figure \ref{fig:low_rank_analysis} (a), (b), (c)). Since singular values indicate the significance of specific features or directions in the data, large singular value changes signify a model's high sensitivity to perturbations in those directions. Adversarial training, designed to enhance model robustness against such perturbations, appears to encourage resistance to changes in these sensitive directions, resulting in a more concentrated distribution of singular value changes around zero.
Another intriguing finding is that CUB-200 exhibits a lower-rank compared to ImageNet, a characteristic that may increase vulnerability to the low-rank attacks. This can be attributed to CUB-200 being a fine-grained dataset focused exclusively on bird classification. Due to this limited variability, models trained on CUB-200 tend to rely on a narrower set of features for classification. Consequently, adversarial perturbations can be more homogeneous, as they only need to manipulate fewer features to achieve misclassification. As we will show in Section \ref{sec:exp}, this explains our ability to craft more effective low-rank adversarial examples on CUB-200.

To further validate the low-rank property of adversarial attacks, we reconstructed adversarial perturbations by retaining only the top 20\% of singular values and vectors. We conducted experiments using the C\&W attack on 1,000 randomly selected images. The results indicate that the median $l_2$ norm of the original C\&W attack is 2.27, while the attacks reconstructed with only the top 20\% of singular values have a median $l_2$ norm of 2.44, which is notably close to that of the original attacks. This finding further supports the low-rank structure of adversarial perturbations.

\bsub{Low Rank Adversarial Attacks. } 
Discovering the low-rank property of adversarial perturbations has significant implications for adversarial ML research. 
From an attack perspective, imposing low-rank constraints on perturbations can significantly improve the efficiency of adversarial example generation, which is particularly critical for black-box attacks, where state-of-the-art methods often require a prohibitively large number of queries to succeed \cite{cheng2019sign,chen2020hopskipjumpattack,brendel2018decision}. 
Such high query volumes lead to considerable commercial costs when targeting commercial APIs. Frequent queries within a short timeframe can also trigger alerts from the service provider’s intrusion detection systems \cite{chen2020hopskipjumpattack,suya2020hybrid}. Furthermore, adversarial examples generated through black-box attacks also tend to be more perceptible than those created by state-of-the-art white-box methods \cite{brendel2018decision}, making them more prone to detection. 
Understanding the low-rank structure in adversarial perturbations enables us to integrate sophisticated gradient estimation methods from existing attacks \cite{chen2020hopskipjumpattack,Vo2022} with low-rank constraints to substantially improve both the effectiveness and efficiency of adversarial example generation. While Esmaeili et al. \cite{esmaeili2023low} has demonstrated improvements in query efficiency, their method is limited by its focus on the low-rank structure in input data, rather than the low-rank structure of adversarial perturbations. Consequently, their optimization not only operates in high-dimensional perturbation space but also fails to effectively integrate with sophisticated adversarial optimization techniques, both resulting in suboptimal solutions.
To illustrate this, we compare our attack with \cite{esmaeili2023low} under the same settings using an EfficientNet model trained on the ImageNet dataset, the detailed settings and results are given in Appendix \ref{appendix:comp_baselines}. Since \cite{esmaeili2023low} operates in untargeted decision-based settings, we restricted our comparison to this context, though our attacks could work in a broader context. Our approach achieves significantly lower $l_2$ distortion than \cite{esmaeili2023low} (1.40 vs. 16.33), highlighting the advantages of exploiting the inherent low-rank nature of adversarial perturbations.

\bsub{Low Rank Adversarial Defenses. } 
From a defense perspective, the low-rank property can significantly enhance the efficiency of adversarial training. Since adversarial training typically involves solving a min-max problem by first crafting adversarial examples and then minimizing model loss, the effectiveness and efficiency of adversarial training are closely tied to the adversarial example generation process.

Specifically, adversarial training aims to optimize the following objective:
\begin{equation}
   \min_{\theta} \mathbb{E}_{(x,y)\sim D} \left[ \max_{\delta \in B(x,\epsilon)} L_{ce}(x + \delta, y) \right],
\end{equation}

where $(x,y) \sim D$ represents the training data, $B(x,\epsilon)$ is the allowable perturbation set, generated with sample $x$. This process imposes a considerable computational and memory overhead compared to standard training. For instance, in the representative PGD adversarial training \cite{madry2018towards}, if the input size is $(m,n,c)$ and the batch size is $B$, the additional adversarial example generation process incurs $3Bmnc$ additional optimizer-stage memory. As shown in Eq. \ref{eq:iterative_defense}, every optimization step is inherently low-rank for PGD attack, thus, we can approximate the adversarial perturbation using two low-rank matrices with size $(m,r,c)$ and $(r,n,c)$, the memory overhead can be reduced to $3B(m+n)rc$. Assuming $m=n$, the memory difference would be $3Bcm(m-2r)$. Given that $r \ll \min(m, n)$, the memory savings scale linearly with $Bm^2$.
This memory reduction is particularly beneficial in the context of large-batch training, which is becoming a common paradigm \cite{liu2022concurrent} and is especially advantageous when training large-scale models, such as diffusion models with high-resolution inputs \cite{rombach2022high} and large language models with high context lengths \cite{MosaicML2023Introducing}, which have become the standard for large-scale models. Additionally, analyzing the distribution of singular values can offer valuable insights into the adversarial vulnerability of models, guiding the development of more robust defenses.
In this paper, we primarily focus on leveraging insights from the low-rank property to craft more effective and efficient adversarial examples in the black-box setting. A systematic exploration of other applications is left for future work.


\section{Threat Model} \label{sec:threat_model}
We now introduce the threat model.
For victim models, we follow the common settings of existing work on black-box adversarial attacks. Specifically, we consider two settings: 
the score-based setting where the adversary could receive confidence scores, and the decision-based setting where the adversary can only receive the predicted label. 

Besides the access to the victim models, we further consider the adversary capable of collecting an auxiliary dataset. 
Unlike many transfer-based attacks \cite{qin2022boosting, wangunified} that assume adversaries have access to high-quality auxiliary datasets (\textit{e.g.}, drawn from the victim models’ training data), our approach significantly relaxes these requirements. Specifically, our approach allows the auxiliary datasets to come from the same modality (\textit{e.g.}, both images) but with substantially different feature and label distributions from the victim datasets. For instance, in our experiments, we use a facial image dataset to attack models trained for bird species classification.
Since the adversary does not need to collect datasets similar to the victim model's training data, they can reasonably collect auxiliary data from diverse internet sources, including publicly available datasets or social media platforms. In scenarios where the adversary is aware of the victim model’s task (\textit{e.g.}, facial recognition), they may strategically choose more relevant datasets (\textit{e.g.}, facial images) to increase attack effectiveness. The auxiliary datasets are also limited in size (\textit{e.g.}, as few as 1,000 samples). 
Additionally, we consider the adversary has access to some pre-trained models, which have a different architecture and training distribution from the victim models. Pre-trained models among multiple domains are readily available online, despite their training data not always being accessible. Platforms like Pytorch Hub \cite{pytorchhub} and TensorFlow Hub \cite{tensorflowhub} offer a set of readily available pre-trained models spanning vision, audio, NLP, etc. These models are not only widely accessible but also easy to deploy.

\section{Method} \label{sec:method}

\subsection{Overview}

\bsub{Design Intuition: } 
Crafting adversarial examples for a ML model involves determining the minimal perturbation needed to shift an input into the model's desired decision region. This perturbation can be considered as a vector in the input space, and determining its direction is crucial. A perturbation in the right direction can move the input to the target decision region with minimal change. In white-box attacks, this direction is derived directly from gradient information, while in black-box attacks, it is estimated through numerous iterative queries and model feedback. However, due to the limited information available in black-box scenarios, these estimates can be imprecise. To effectively facilitate the estimation of perturbation, we propose leveraging the insight that adversarial perturbations typically reside within a low-rank subspace. Then by restricting the perturbation direction to the low-rank subspace, we can improve the estimation accuracy and efficiency, thereby enabling us to develop more effective and efficient adversarial attacks.

\begin{figure}[t]
    \centering
    \includegraphics[scale=0.43]{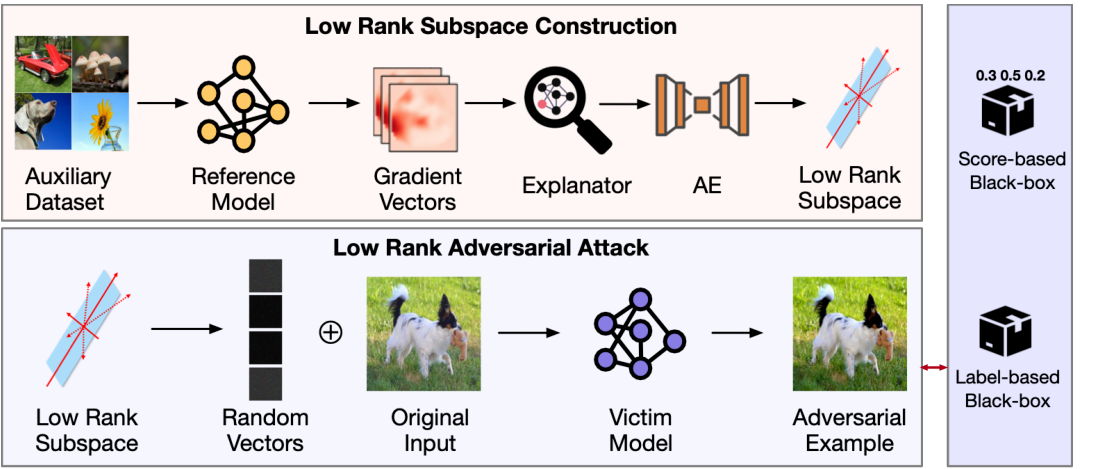}    \caption{Overview of the low-rank adversarial attack.} \label{fig:overall_pipeline}
\end{figure}

\bsub{Challenges: } Leveraging the low-rank property to craft efficient and effective adversarial attacks presents significant challenges. Particularly, two primary challenges stand out:

\vspace{2pt}\noindent\textit{C1. Constructing the Low-rank Subspace in Black-box Setting: } In a black-box setting, the adversary lacks direct access to the gradients used to update the adversarial example, making it impossible to directly decompose these updates into a low-rank matrix as outlined in Eq. \ref{eq1:low_rank}. This limitation introduces significant difficulties in constructing the low-rank subspace necessary for adversarial example generation, necessitating the development of novel methods.

\vspace{2pt}\noindent\textit{C2. Utilizing the Low-rank Subspace for Adversarial Attacks: } The second challenge lies in effectively utilizing the constructed low-rank subspace to perform adversarial attacks. Given the sophistication of existing research on black-box attacks, we aim to incorporate the low-rank subspace into these attacks to further enhance their effectiveness and efficiency. The main difficulty arises from the varied mechanisms employed by different attacks, raising the challenge of developing a generalizable framework that can be applied effectively across diverse attack strategies.

\bsub{Overview: }
In this paper, we propose a novel low-rank adversarial attack framework that fully leverages the low-rank property to enhance both the efficiency and effectiveness of the attack. Figure \ref{fig:overall_pipeline} provides an overview of our attack framework. To address challenge C1, we construct a shadow subspace by leveraging an auxiliary dataset and a reference model to generate gradient vectors. By applying explainable machine learning techniques, we identify the critical components of these gradients. These key elements are then used to train an autoencoder, which constructs a low-rank subspace. To address challenge C2, we first abstract various black-box attack methods and identify integration points for incorporating the low-rank subspace into existing attack strategies. A common characteristic of these attacks is their reliance on random vectors for gradient estimation. By confining these random vectors within the low-rank subspace, we restrict the adversarial optimization to the low-rank space. The details of the low-rank subspace construction and the low-rank adversarial attack are further elaborated in Sections \ref{subsec:subspace_optim}, and \ref{subsec:attack_frame}, respectively.

\subsection{Low Rank Subspace Construction} \label{subsec:subspace_optim}

In a black-box setting, the lack of access to the gradient information of the victim model makes it impossible to directly construct the low-rank subspace. However, inspired by findings that the decision boundaries of different models exhibit certain similarities and correlations \cite{liu2022delving,dong2018boosting}, we hypothesize that the low-rank subspaces of different models may also share similar properties. As shown in our experiments in Section \ref{sec:exp}, we observe that the low-rank subspaces demonstrate a high degree of transferability across different model architectures and datasets. This transferability likely stems from the fact that models trained on natural images tend to learn similar low-level features regardless of their specific domain \cite{yosinski2014transferable,zeiler2014visualizing}, which provide a common space where adversarial perturbations can effectively operate. This enables us to leverage an auxiliary dataset, one that does not share the same data distribution or labels as the victim model's training data, along with a publicly available pre-trained model of a different architecture and training distribution to construct a transferable low-rank subspace.




To reconstruct the low-rank subspace, we introduce a subspace denoted as $\mathbb{V}^{r \times r}$, where $\mathbb{V}$ is a subset of the original space $\mathbb{O}^{m \times n}$. Any vector $\upsilon \in \mathbb{V}^{r \times r}$ represents a low-dimensional search direction, while $\theta \in \mathbb{O}^{m \times n}$ represents the search direction in the full-dimensional space. It is important to note that $r \ll \min(m, n)$, making gradient estimation in the low-rank subspace more efficient than in the original space. The effectiveness of adversarial attacks is closely tied to selecting an appropriate subspace $\mathbb{V}^{r \times r}$. To construct this subspace, we compute gradients on an auxiliary dataset using a reference model. However, the gradients may be noisy and exhibit sharp fluctuations in local regions due to the piecewise differentiable nature of activation functions (\textit{e.g.}, ReLU) \cite{smilkov2017smoothgrad}. Consequently, accurately capturing the transferable subspace becomes challenging.

To address this, we leverage explainable machine learning techniques. Explainable ML has gained significant attention in recent years, often used to generate attribution maps that highlight the contribution of individual features to the model's decision-making process \cite{shrikumar2017learning,lundberg2017unified,selvaraju2017grad}. By first generating attribution maps and subsequently multiplying the gradients by these maps, we identify the gradients that are regarded as crucial for model decisions. This allows us to construct a more accurate and refined subspace.


There are several methods for constructing low-rank subspaces.
One common approach is to use Principal Component Analysis (PCA) \cite{wold1987principal}, however, PCA is a linear method, which may not be able to capture the complex, non-linear patterns of gradient information. 
Also, PCA suffers computation efficiency and memory scalability issues \cite{elgamal2015spca} on large high-dimensional datasets. 
To address this limitation, we propose an autoencoder model to learn the low-dimensional space effectively. Autoencoder (AE) is an unsupervised machine learning model, and it has been widely used for feature learning or dimensionality reduction \cite{bank2023autoencoders}. A typical AE consists of two parts, an encoder and a decoder. An AE maps input data $x$ to a reduced feature space through the encoder $z=e(x)$ and subsequently reconstructs the original data from this reduced representation via the decoder function $r=d(z)$. For the design of the AE, we adopt the U-net architecture \cite{ronneberger2015u}, with detailed specifications provided in Appendix \ref{appendix:autoencoder}.
We propose the following loss function to optimize the AE:

\begin{equation}
    l(g,m,\hat{g},\hat{m}) = \frac{1}{N}\sum_{i=1}^{N} (g_i-\hat{g_i})^2 + \alpha \cdot (m_i-\hat{m_i})^2,
\end{equation}

where the $g$ represent the gradient, $m$ represents the masked gradient, $\hat{g}$ represents the reconstructed gradient, and $\hat{m}$ represent the reconstructed masked gradient.

\subsection{Low Rank Adversarial Attack} \label{subsec:attack_frame}


After constructing the low-rank subspace, the next step is to utilize it for performing adversarial attacks. Given the sophistication of existing research on adversarial attacks, we aim to integrate the low-rank subspace with the existing advanced adversarial attacks. To achieve this, we begin by abstracting different black-box attacks and identifying the points of integration.


In score-based attacks, adversarial attacks generally start from the original seed image and gradually search for better perturbations with the estimated gradients. In hard-label settings, due to the limited information, attacks usually begin with a reference image that satisfies the attacker's objective and then focus on minimizing the perturbation size with estimated gradients. Despite the differences in the underlying mechanisms of various attacks, they generally rely on a random vector to perturb input for gradient estimation. Score-based attacks leverage such vectors to estimate the gradient by observing the change in output scores for slightly perturbed versions of the input \cite{ilyas2018prior,chen2017zoo,uesato2018adversarial}. Conversely, hard-label attacks use these vectors to approximate the model's decision boundary by observing how perturbations affect the predicted label \cite{cheng2018query,cheng2019sign,chen2020hopskipjumpattack}. Therefore, we can restrict the adversarial optimization within the low rank subspace by confining the random vectors to this space, facilitating the optimization.

%
The algorithm for our low-rank adversarial attack is outlined in Alg. \ref{alg:adversarial_framework}. The core idea of our approach involves sampling random vectors and performing gradient estimation within a low-dimensional subspace (lines 3-4). Adversaries first use gradients and explanations derived from the auxiliary dataset and reference model to train an autoencoder (AE). During the optimization phase, they sample random vectors from a low-dimensional subspace, denoted as $\mathbb{V}^{r \times r}$. These vectors are then remapped to the original gradient space, $\mathbb{O}^{m \times n}$, using the previously trained decoder function $d$. Finally, the adversaries employ these remapped vectors to estimate gradients, thereby enhancing the attack efficacy. 


\begin{algorithm}
\caption{Low Rank Adversarial Attack} \label{alg:adversarial_framework}
\begin{algorithmic}[1]

\Require Original example $x$, target model $M$, salient mask $m$, maximum iteration $T$, number of random sampling $N$, decoder model $d$, low-rank space $\mathbb{V}^{r \times r}$ and original space $\mathbb{O}^{m \times n}$. 
\Ensure The adversarial example $x^{*}$.
 \State $x_{adv} \gets x_0$;
\For{$t = 1,2,...,T$}



    \State Sample random vectors $\upsilon_1,...,\upsilon_N \in \mathbb{V}^{r \times r}$;    
    \State $\mu_1,\mu_2,...,\mu_N$ = $d(\upsilon_1),d(\upsilon_2),...,d(\upsilon_N)$;
    
    \State $\delta \gets$ estimate\_perturbation($x_{adv}$, $M$, $\mu_1,...,\mu_N$)
    \State $x_{adv} \gets$ apply\_perturbation($x$, $\delta$)

    
\EndFor
\State \Return $x_{adv}$

\end{algorithmic}

\end{algorithm}


\section{Experiment} \label{sec:exp}

\subsection{Experimental Setup} \label{subsec:exp_setup}

\bsub{Datasets and Models. }
We conduct our experiments on three widely used benchmark datasets: ImageNet \cite{deng2009imagenet}, CUB-200-2011 \cite{wah2011caltech}, and Stanford Cars \cite{krause20133d}. Details of each dataset are provided in Appendix~\ref{appendix:exp_setting}. To evaluate our attack’s effectiveness across different model architectures, we use Vision Transformer (ViT) \cite{dosovitskiy2020image} and EfficientNet (EffNet) \cite{tan2019efficientnet} on ImageNet, ResNet-50 \cite{he2016deep} on CUB-200-2011, and ResNet-34 \cite{he2016deep} on Stanford Cars.  
Following existing adversarial attack studies \cite{chen2020hopskipjumpattack,fu2022autoda,metzen2017detecting,wan2024bounceattack}, we select 100 correctly classified test images for each dataset and model for computational efficiency. These images are randomly drawn from the respective test sets and evenly distributed across 20 classes. To further evaluate the generalizability of our attack, we additionally sample 1,000 images from ImageNet when using Vision Transformer for all attacks. For all experiments, we clip the perturbed image into the input domain $[0, 1]$ by default.

\bsub{Evaluation Metrics. }
Following the state-of-the-art studies \cite{chen2020hopskipjumpattack,cheng2019sign,Vo2022}, we use the minimum $l_p$ ($p\in{(2,\infty)}$) distortion w.r.t the original example obtained during the adversarial attacks as the evaluation metric. For attacks that find the minimum perturbations, the minimum distortion can be efficiently calculated. For attacks that craft adversarial examples with constrained perturbations $\epsilon$ (\textit{e.g.}, Bandits \cite{ilyas2018prior}), we perform a binary search on $\epsilon$ to find its minimum value that enables successful attacks. We report the median $l_p$ distortions across all attack images following \cite{chen2020hopskipjumpattack,brendel2018decision,cheng2019sign}. We also report the robust accuracy of the model at given perturbation budget $\epsilon$, which is calculated by counting the proportion of adversarial examples whose minimum distortion magnitude is larger than $\epsilon$.
To further evaluate attack efficiency, we utilize the \textit{Accuracy vs. Query} curve. We also utilize \textit{Accuracy vs. Perturbation Budget} curve to comprehensively evaluate attack effectiveness following \cite{dong2020benchmarking,carlini2019evaluating}.



\begin{table}[t]
\small
\centering
\caption{The median $l_2$ distortion and model accuracy ($\epsilon=2$) of applying our proposed method in Bandits attack.}
\label{tab:score-based}
\renewcommand{\arraystretch}{0.94}
\setlength{\belowrulesep}{1.5pt}
\setlength{\tabcolsep}{1.2pt}
\begin{tabular}{c|cc|cc}
\toprule
\hline
\multirow{2}{*}{\textbf{\begin{tabular}[c]{@{}c@{}}Explanation \\ Method\end{tabular}}} & \multicolumn{2}{c|}{\textbf{CUB}} & \multicolumn{2}{c}{\textbf{ImageNet}} \\ \cline{2-5} 
                                                                                        & Median Dist.      & Accuracy      & Median Dist.        & Accuracy        \\ \hline
Naive                                                                                   & 11.18             & 0.94          & 10.55               & 0.98            \\ \hline
GradCAM                                                                                 & 2.73              & 0.68          & 4.00                & 0.84            \\ \hline
InvGrad                                                                                 & 2.70              & 0.67          & 3.97                & 0.83            \\ \hline
LIME                                                                                    & 2.74              & 0.68          & 4.01                & 0.84            \\ \hline
SHAP                                                                                    & 2.73              & 0.68          & 4.07                & 0.86            \\ \hline
\end{tabular}
\end{table}

\begin{table*}[t]
\small
\centering
\caption{The median $l_2$ distortion and model accuracy ($\epsilon=2$ for untargeted attack, $\epsilon=10$ for targeted attack) of adversarial attacks on different attack settings, datasets, and explanations under the decision-based black-box setting.}
\label{tab:hard_label_comprehensive}
\renewcommand{\arraystretch}{0.99}
\setlength{\belowrulesep}{1.5pt}
\setlength{\tabcolsep}{1.5pt}
\begin{tabular}{c|c|c|cc|cc|cc|cc|cc}
\toprule
\hline
\multirow{2}{*}{\textbf{Attack Setting}} & \multirow{2}{*}{\textbf{\begin{tabular}[c]{@{}c@{}}Attack \\ Method\end{tabular}}}                                  & \multirow{2}{*}{\textbf{Data/Model}} & \multicolumn{2}{c|}{\textbf{Naive HSJA}} & \multicolumn{2}{c|}{\textbf{GradCAM}} & \multicolumn{2}{c|}{\textbf{InvGrad}} & \multicolumn{2}{c|}{\textbf{LIME}} & \multicolumn{2}{c}{\textbf{SHAP}} \\ \cline{4-13} 
                                         &                                                                          &                                      & Med. Dist.          & Acc.         & Med. Dist.        & Acc.        & Med. Dist.        & Acc.        & Med. Dist.       & Acc.      & Med. Dist.      & Acc.      \\ \hline
\multirow{16}{*}{Pre-trained}            & \multirow{4}{*}{HSJA}                                                    & ImageNet/ViT                         & 1.63                  & 0.39             & 1.07                & 0.20            & 1.12                & 0.21            & 1.07               & 0.20          & 1.07              & 0.20          \\
                                         &                                                                          & ImageNet/EffNet                      & 2.05                  & 0.51             & 1.40                & 0.31            & 1.30                & 0.26            & 1.49               & 0.27          & 1.39              & 0.32          \\
                                         &                                                                          & CUB/ResNet-50                        & 2.64                  & 0.66             & 0.82                & 0.20            & 0.85                & 0.19            & 0.86               & 0.20          & 0.83              & 0.17          \\
                                         &                                                                          & Stanford/ResNet-34                   & 2.94                  & 0.73             & 2.51                & 0.59            & 2.42                & 0.58            & 2.48               & 0.59          & 2.40              & 0.59          \\ \cline{2-13} 
                                         & \multirow{4}{*}{Sign-Opt}                                                & ImageNet/ViT                         & 5.57                  & 0.75             & 1.71                & 0.46            & 1.65                & 0.46            & 1.65               & 0.46          & 1.74              & 0.47          \\
                                         &                                                                          & ImageNet/EffNet                      & 5.32                  & 0.86             & 2.46                & 0.57            & 2.57                & 0.60            & 2.56               & 0.59          & 2.57              & 0.59          \\
                                         &                                                                          & CUB/ResNet-50                        & 3.05                  & 0.67             & 0.55                & 0.07            & 0.55                & 0.06            & 0.56               & 0.07          & 0.58              & 0.07          \\
                                         &                                                                          & Stanford/ResNet-34                   & 3.09                  & 0.67             & 2.23                & 0.60            & 2.27                & 0.60            & 2.27               & 0.61          & 2.27              & 0.61          \\ \cline{2-13} 
                                         & \multirow{4}{*}{RamBoAttack}                                             & ImageNet/ViT                         & 1.88                  & 0.46             & 1.10                & 0.19            & 1.07                & 0.19            & 1.06               & 0.19          & 1.07              & 0.20          \\
                                         &                                                                          & ImageNet/EffNet                      & 2.05                  & 0.54             & 1.32                & 0.26            & 1.34                & 0.27            & 1.34               & 0.27          & 1.32              & 0.27          \\
                                         &                                                                          & CUB/ResNet-50                        & 2.53                  & 0.60             & 1.04                & 0.28            & 1.04                & 0.28            & 1.05               & 0.28          & 1.04              & 0.29          \\
                                         &                                                                          & Stanford/ResNet-34                   & 2.73                  & 0.66             & 1.79                & 0.42            & 1.81                & 0.43            & 1.79               & 0.42          & 1.79              & 0.43          \\ \cline{2-13} 
                                         & \multirow{4}{*}{\begin{tabular}[c]{@{}c@{}}HSJA\\ Targeted\end{tabular}} & ImageNet/ViT                         & 4.14                  & 0.06             & 2.78                & 0.01            & 2.75                & 0.01            & 2.83               & 0.01          & 2.83              & 0.00          \\
                                         &                                                                          & ImageNet/EffNet                      & 6.50                  & 0.28             & 5.84                & 0.24            & 5.77                & 0.17            & 6.10               & 0.17          & 5.68              & 0.20          \\
                                         &                                                                          & CUB/ResNet-50                        & 9.14                  & 0.43             & 3.64                & 0.01            & 3.48                & 0.01            & 3.73               & 0.01          & 3.47              & 0.02          \\
                                         &                                                                          & Stanford/ResNet-34                   & 9.62                  & 0.44             & 6.63                & 0.08            & 6.45                & 0.07            & 6.61               & 0.08          & 6.64              & 0.08          \\ \hline
\multirow{4}{*}{Non-pre-trained}         & \multirow{4}{*}{HSJA}                                                    & ImageNet/ViT                         & 1.63                  & 0.39             & 1.35                & 0.30            & 1.31                & 0.29            & 1.32               & 0.28          & 1.35              & 0.30          \\
                                         &                                                                          & ImageNet/EffNet                      & 2.05                  & 0.51             & 1.36                & 0.25            & 1.36                & 0.28            & 1.45               & 0.30          & 1.35              & 0.24          \\
                                         &                                                                          & CUB/ResNet-50                        & 1.81                  & 0.45             & 0.58                & 0.06            & 0.56                & 0.10            & 0.58               & 0.10          & 0.62              & 0.10          \\
                                         &                                                                          & Stanford/ResNet-34                   & 3.84                  & 0.84             & 3.40                & 0.76            & 3.34                & 0.77            & 3.30               & 0.76          & 3.46              & 0.75          \\ \hline
\end{tabular}
\end{table*}

\bsub{Adversarial Attacks. }
We select representative attacks for comparison. 
Since decision-based attacks are considered to have the most practical threat models \cite{fu2022autoda,cheng2018query}, our evaluation mainly focuses on this case. 
Specifically, we choose the Bandits Attack \cite{ilyas2018prior} for score-based attack and select three decision-based attacks: HopSkipJumpAttack (HSJA) \cite{chen2020hopskipjumpattack}, Sign-Opt Attack \cite{cheng2019sign}, and RamBoAttack \cite{Vo2022}, which are widely recognized in the adversarial literature for their effectiveness and efficiency. To ensure a fair comparison, we use the same attack configurations and parameters for both the naive attacks and the low-rank attacks. Detailed configurations are given in Appendix \ref{appendix:exp_setting}. We run each attack with sufficient queries to ensure it is approaching convergence.


\bsub{Explanation Methods.}
Since our approach utilizes explanation methods to identify critical gradient components, we evaluate its robustness across different explanation techniques. Specifically, we select four representative explanators: IG~\cite{sundararajan2017axiomatic}, GradCAM~\cite{selvaraju2017grad}, SHAP \cite{lundberg2017unified}, and LIME \cite{ribeiro2016should}. 


\bsub{Victim Model Training. } 
We consider two training settings for the victim models: pre-trained and non-pre-trained. In computer vision, it has become the widely accepted standard to first train a model on the ImageNet dataset and then adapt the pre-trained model for a new target task \cite{huh2016makes}. This pre-training paradigm has demonstrated effectiveness across various applications. In this study, we examine whether such pre-training influences the effectiveness of adversarial attacks by comparing attacks on both pre-trained and non-pre-trained models.
For the pre-trained models, we adopt the following approaches: For ViT and EfficientNet, we utilize the pre-trained weights available in its official repository. For the other three models, we follow standard practice, training them from models that were pre-trained on the ImageNet dataset. For the non-pre-trained models, we train all models from scratch, without initialization from ImageNet weights. Details of the training configurations for each model are provided in Appendix \ref{appendix:exp_setting}. All models achieved near state-of-the-art accuracy on their respective datasets.

\bsub{Auxiliary Dataset and Reference Model. } 
Our method requires an auxiliary dataset and reference model for subspace construction. For attacking ImageNet, CUB-200, and Stanford Cars, we randomly select 9K images from the ImageNet validation set (ensuring no overlap with images used in attacks) as the auxiliary dataset. In the non-pre-trained setting for attacking ImageNet, we randomly select 9K images from the Caltech-101 dataset. 
We further explore the effects of using Stanford Cars and the Large-scale CelebFaces Attributes (CelebA) dataset as auxiliary datasets and investigate the impacts of varying auxiliary dataset sizes in Section~\ref{subsec:exp_subspace}.
Notably, these datasets can vary significantly in label and feature spaces: ImageNet and Caltech-101 cover a broad range of general objects, whereas CUB-200, Stanford Cars, and CelebA focus on birds, cars, and faces, respectively.
For generating gradients to train the autoencoder, we use DenseNet-121 \cite{huang2017densely} as the reference model, which has a different architecture from the victim models. We thoroughly investigate the impact of auxiliary dataset and reference model choices on attack success in Section \ref{subsec:exp_subspace}.

\begin{figure*}[h]
  \centering
    \begin{minipage}{0.22\textwidth}
    \includegraphics[width=\linewidth]{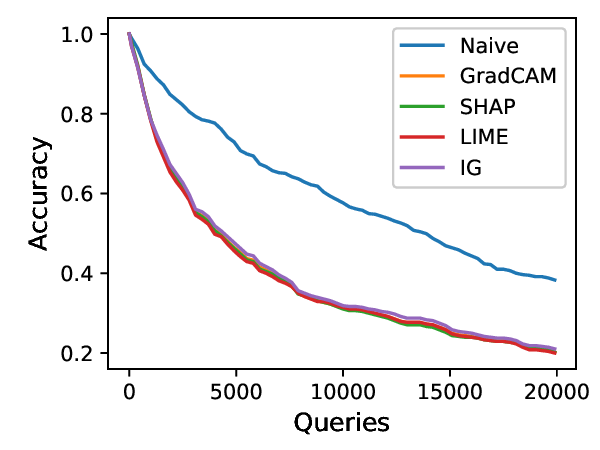}\\
    \includegraphics[width=\linewidth]{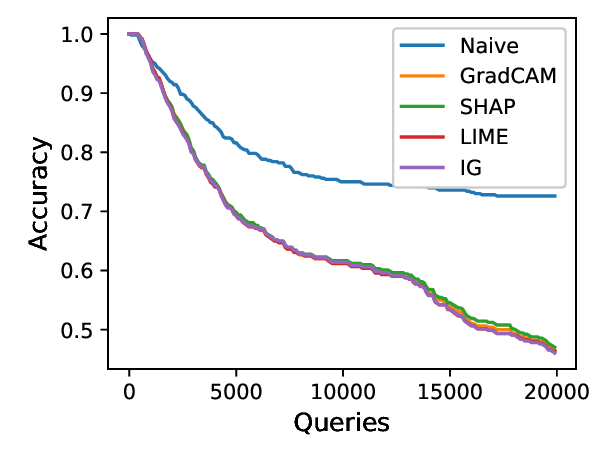}\\
    \includegraphics[width=\linewidth]{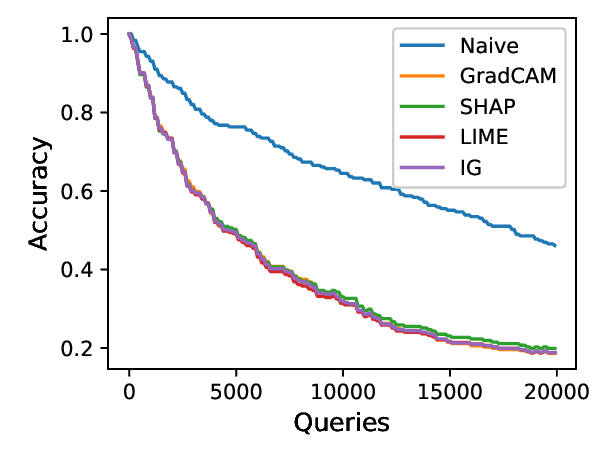}\\
    \includegraphics[width=\linewidth]{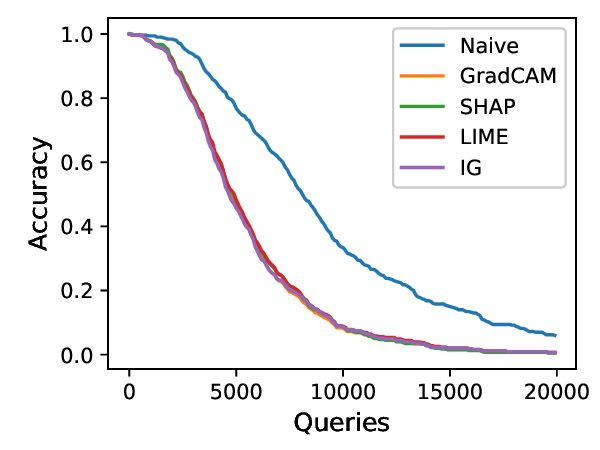}\\
    \includegraphics[width=\linewidth]{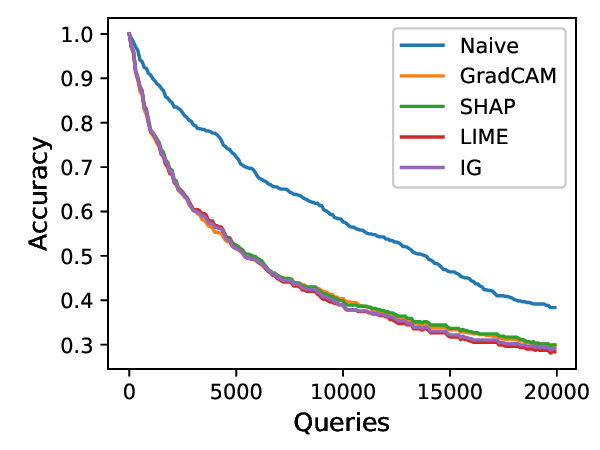}
    \caption*{(a) ImageNet/ViT}
  \end{minipage}
  \hfill
   \begin{minipage}{0.22\textwidth}
   \includegraphics[width=\linewidth]{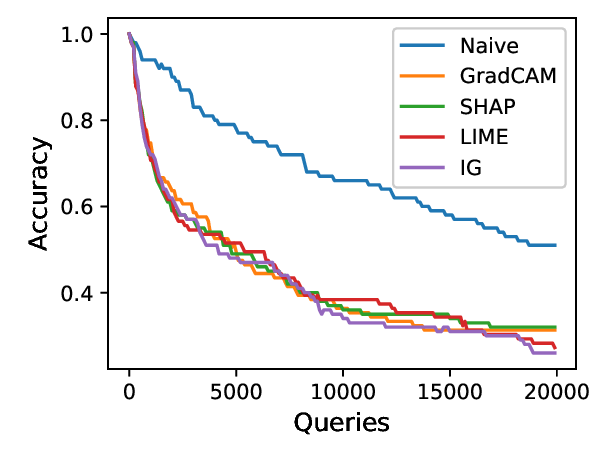}\\
   \includegraphics[width=\linewidth]{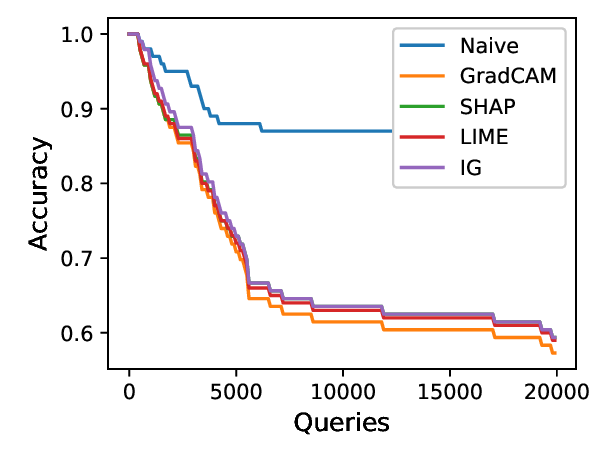}\\
   \includegraphics[width=\linewidth]{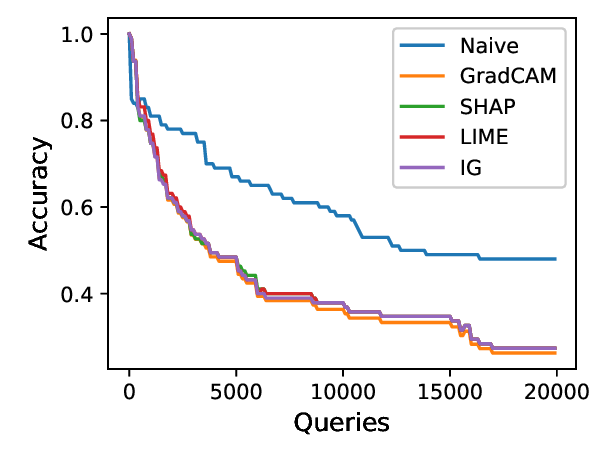}\\
   \includegraphics[width=\linewidth]{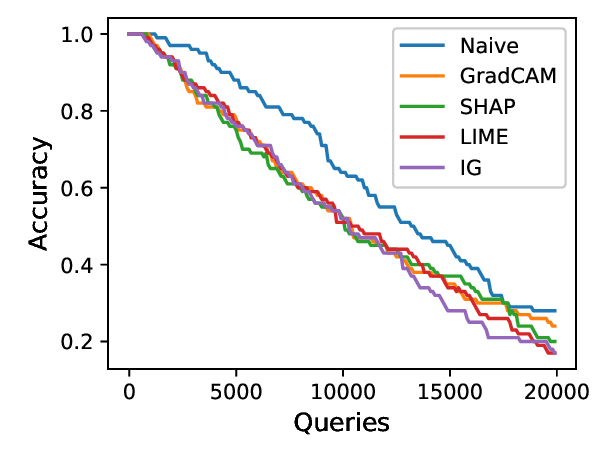}\\
    \includegraphics[width=\linewidth]{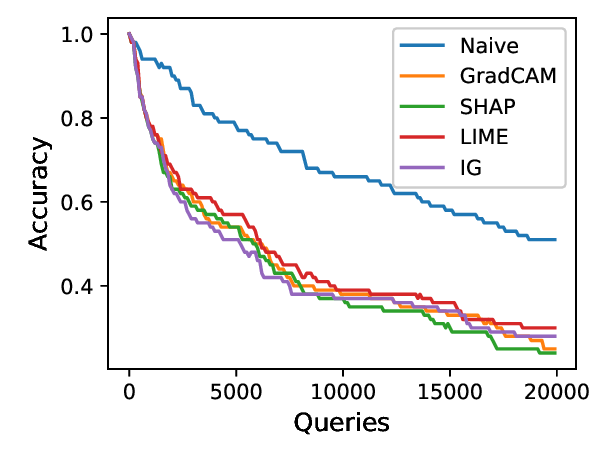}
    \caption*{(b) ImageNet/EffNet}
  \end{minipage}
  \hfill
    \begin{minipage}{0.22\textwidth}
    \includegraphics[width=\linewidth]{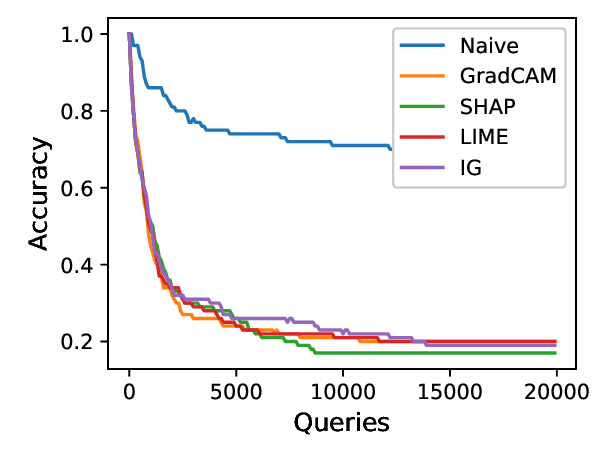}\\
    \includegraphics[width=\linewidth]{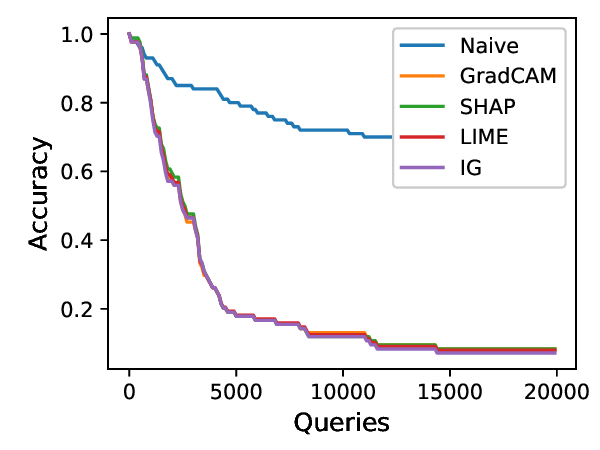}\\
    \includegraphics[width=\linewidth]{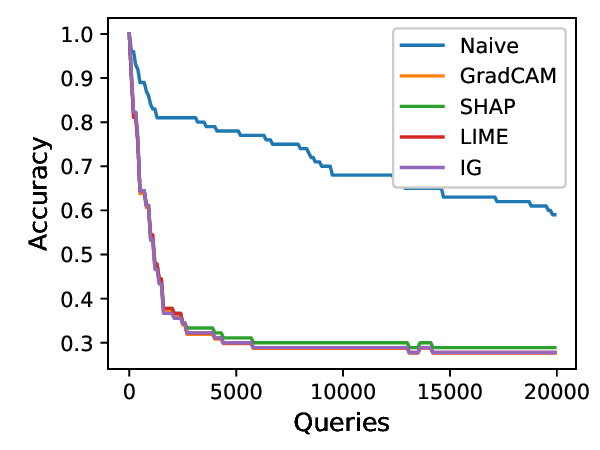}\\
    \includegraphics[width=\linewidth]{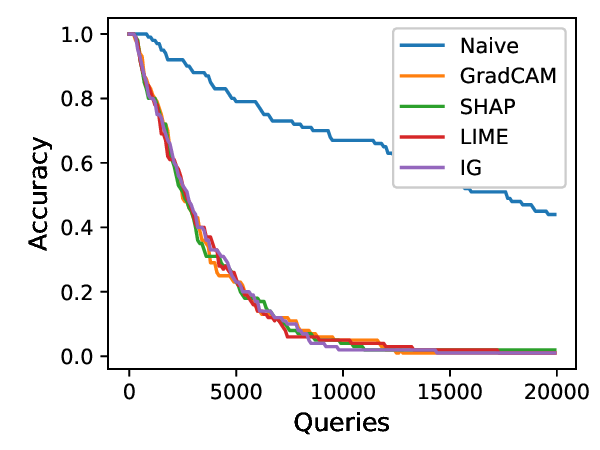}\\
    \includegraphics[width=\linewidth]{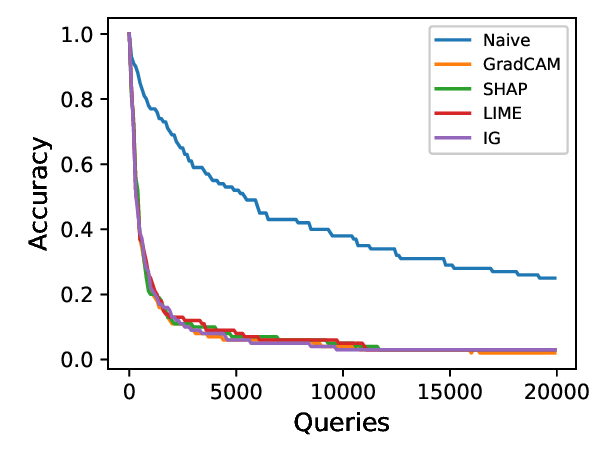}
    \caption*{(c) CUB/ResNet-50}
  \end{minipage}
  \hfill
    \begin{minipage}{0.22\textwidth}
    \includegraphics[width=\linewidth]{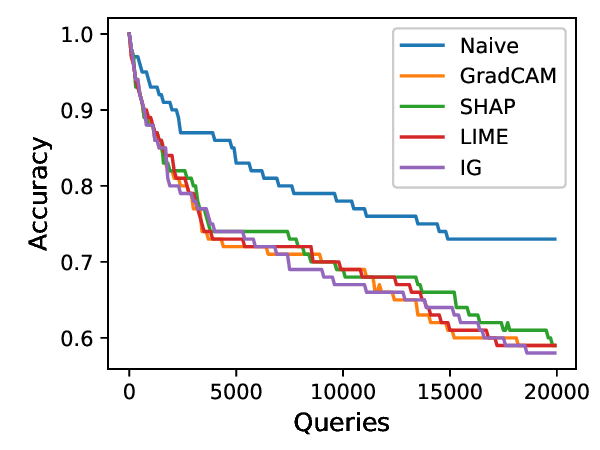}\\
    \includegraphics[width=\linewidth]{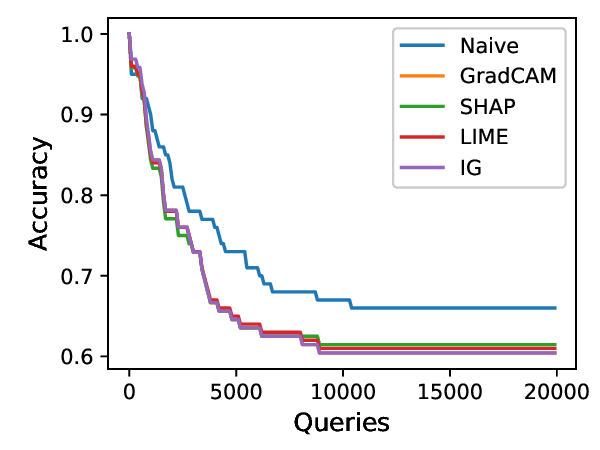}\\
    \includegraphics[width=\linewidth]{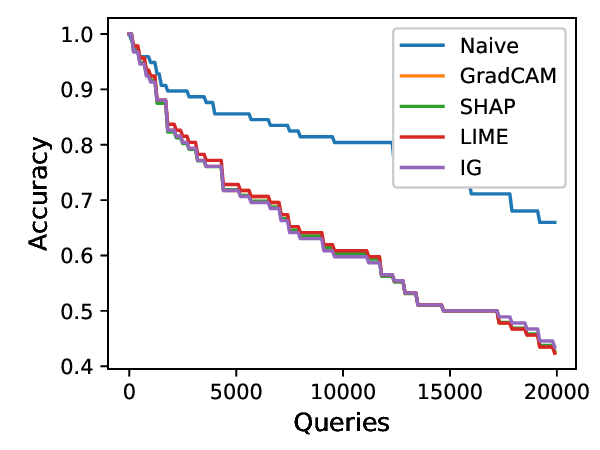}\\
    \includegraphics[width=\linewidth]{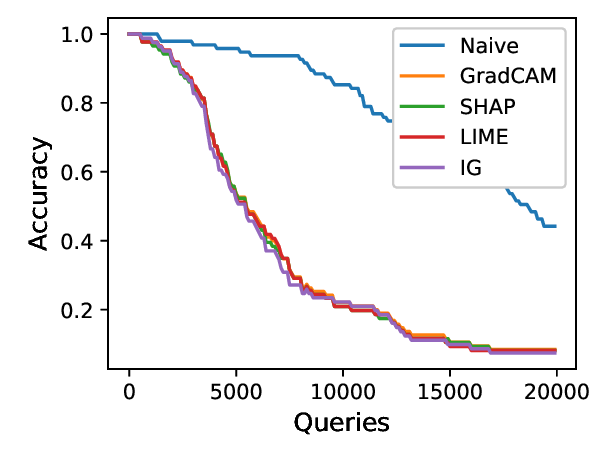}\\
    \includegraphics[width=\linewidth]{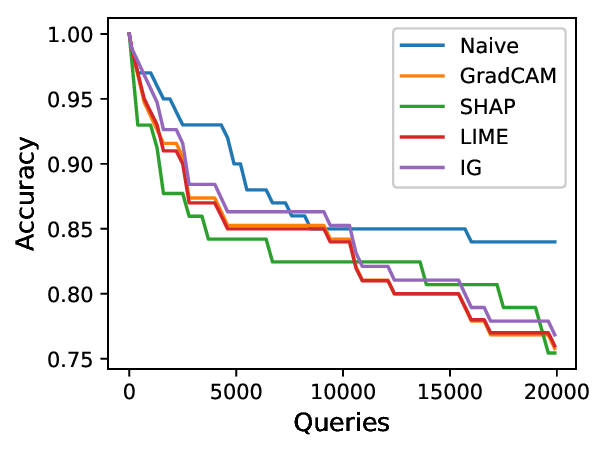}
    \caption*{(d) Stanford/ResNet-34}
  \end{minipage}

  \caption{The accuracy vs. query curves ($\epsilon=2$ for untargeted attack, $\epsilon=10$ for targeted attack) across four datasets and explanators. The first four rows display curves for pre-trained models, while the last row represents non-pre-trained models. The first row shows curves for HSJA, the second row shows Sign-Opt, the third row shows RamBoAttack, and the fourth row highlights the HSJA targeted attack, and the final row shows HSJA curves for non-pre-trained models.}
  \label{fig:enhanced_comprehensive_query}
\end{figure*}

\subsection{Evaluation Results} \label{subsec:eval_res}


%

\subsubsection{Score-based Black-box Setting}

\begin{figure}[h]
\subfigure[CUB]{
\centering
\includegraphics[width=0.465\linewidth]{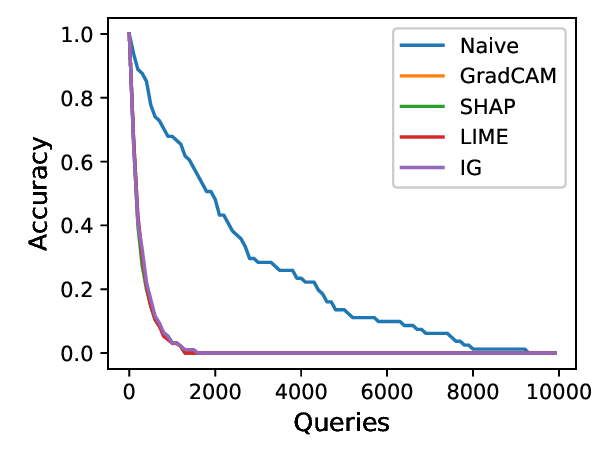}
}
\hfill
\subfigure[ImageNet]{
\centering
\includegraphics[width=0.465\linewidth]{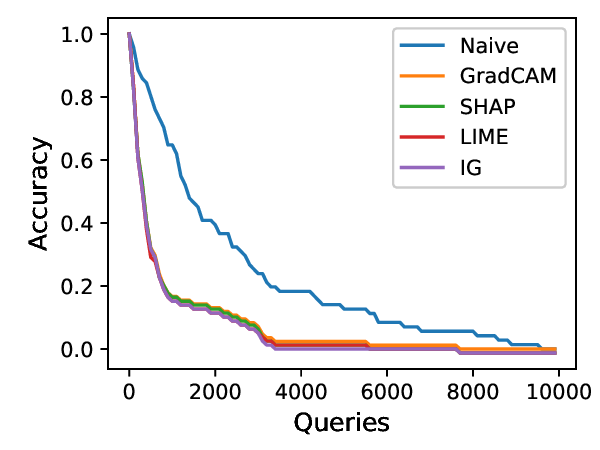}
}
\caption{The accuracy vs. queries curves of applying our method in Bandits attack with different datasets.}
\label{fig:score-based-query}
\end{figure}

In this section, we evaluate the effectiveness of low-rank adversarial attacks in a score-based black-box setting. Our low-rank attack framework is applied to the Bandits Attack method \cite{ilyas2018prior}, using its naive attack variant as a baseline for comparison. The experiments are conducted on the CUB-200 and ImageNet datasets, with results presented in Table \ref{tab:score-based} and Figure \ref{fig:score-based-query}. Our framework significantly enhances the robustness assessment of models across various explanation methods. For example, when incorporating GradCAM explanations into the attacks, the median distortion on the CUB dataset is reduced to 2.73, a substantial improvement over the 11.18 median distortion observed with the naive attack, demonstrating the efficacy of our approach. Moreover, the attack's effectiveness remains consistent across different explanation methods, further illustrating the robustness of our approach. These explanation methods effectively highlight critical gradient components, allowing for the construction of a tight and accurate low-rank subspace, which in turn leads to more effective and efficient adversarial attacks.


\subsubsection{Decision-based Black-box Setting} \label{subsubsec:hard_label}
We mainly focus on assessing the effectiveness of low-rank attacks on the decision-based black-box scenario under both the pre-trained and the non-pre-trained setting. Our goal is to systematically examine how our attack performs under different training settings. While the original training datasets of victim models might follow a distribution distinct from that of the auxiliary dataset, the pre-training approach may inadvertently align with the distribution of the auxiliary dataset. Conversely, in the non-pre-training scenario, victim models are trained from scratch, ensuring the auxiliary dataset diverges completely from the victim model training data distribution.

\begin{figure*}[!h]
  \centering
    \begin{minipage}{0.215\textwidth}
    \includegraphics[width=\linewidth]{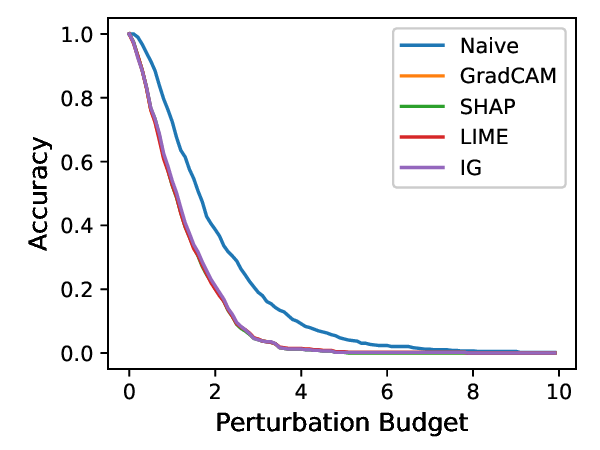}\\
    \includegraphics[width=\linewidth]{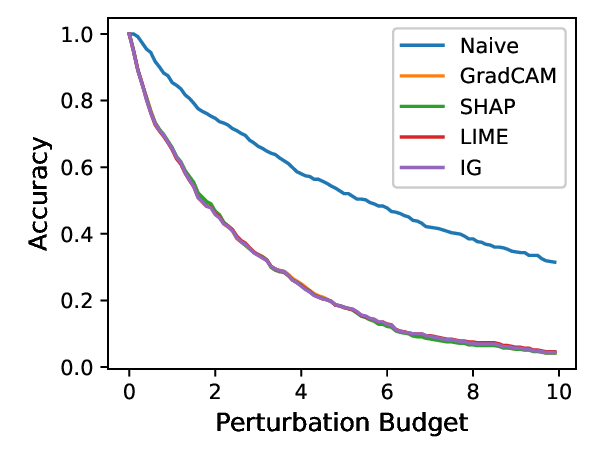}\\
    \includegraphics[width=\linewidth]{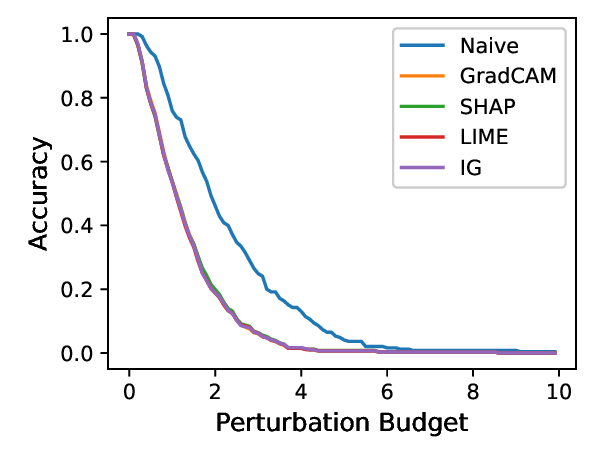}\\
    \includegraphics[width=\linewidth]{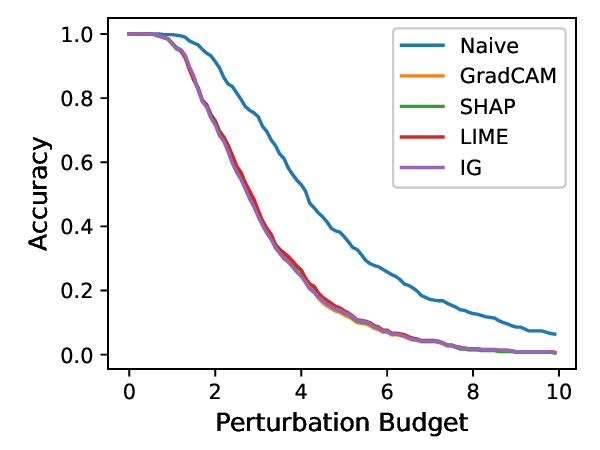}\\
    \includegraphics[width=\linewidth]{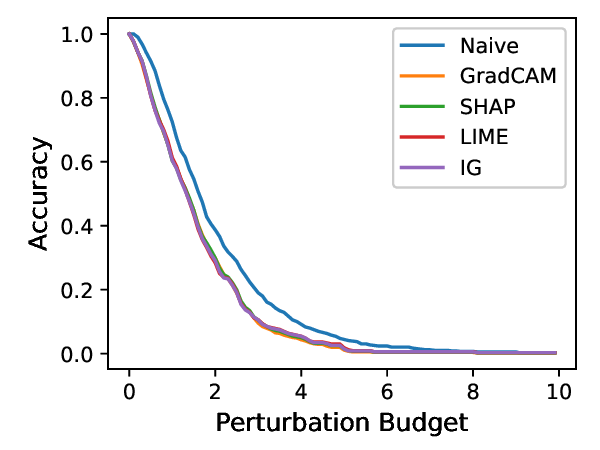}
    \caption*{(a) ImageNet/ViT}
  \end{minipage}
  \hfill
   \begin{minipage}{0.215\textwidth}
   \includegraphics[width=\linewidth]{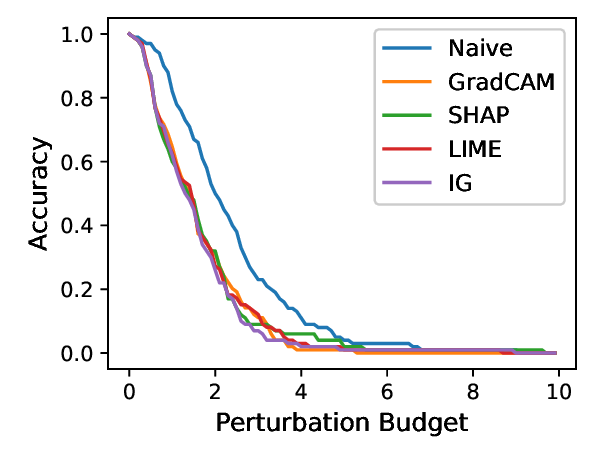}\\
    \includegraphics[width=\linewidth]{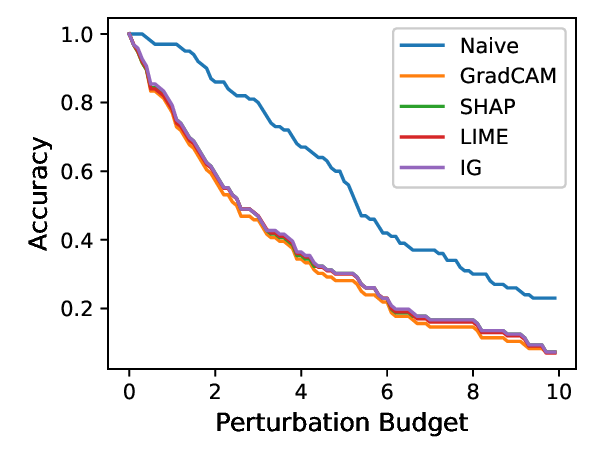}\\
    \includegraphics[width=\linewidth]{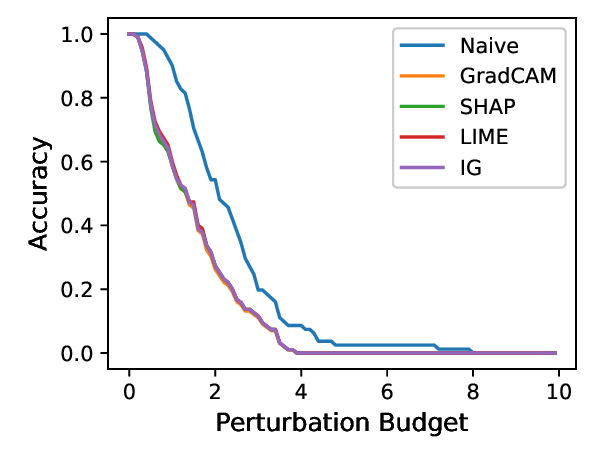}\\
    \includegraphics[width=\linewidth]{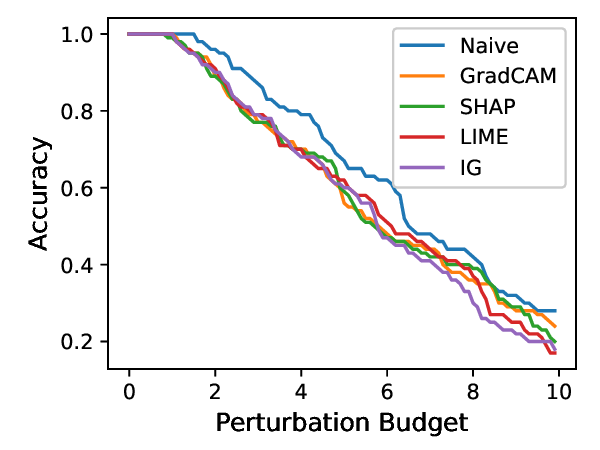}\\
    \includegraphics[width=\linewidth]{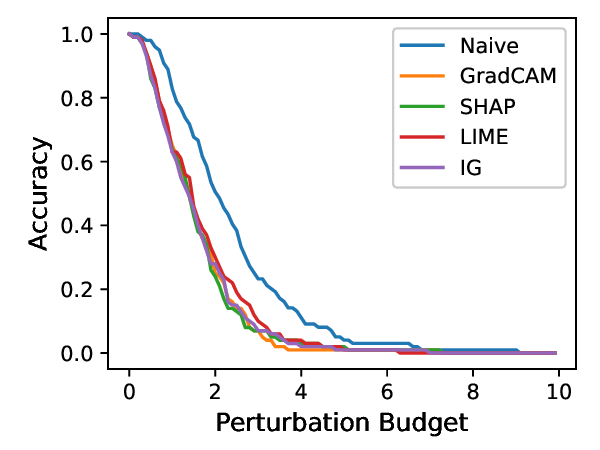}
    \caption*{(b) ImageNet/EffNet}
  \end{minipage}
  \hfill
   \begin{minipage}{0.215\textwidth}
    \includegraphics[width=\linewidth]{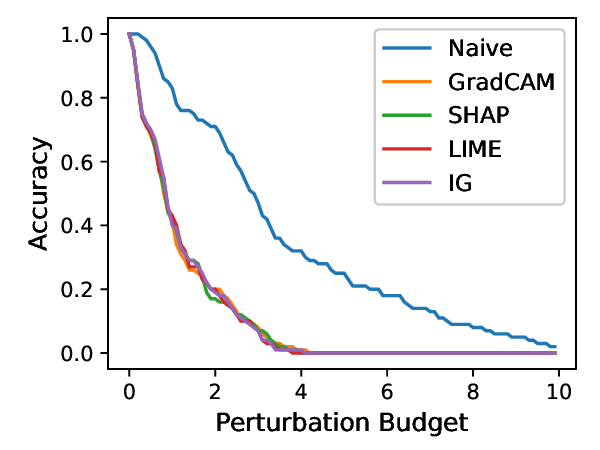}\\
    \includegraphics[width=\linewidth]{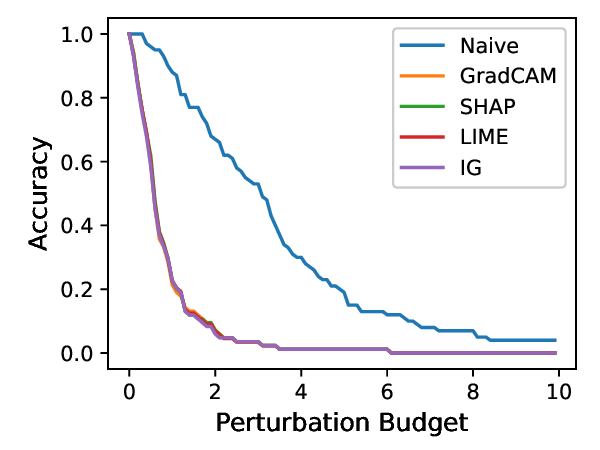}\\
    \includegraphics[width=\linewidth]{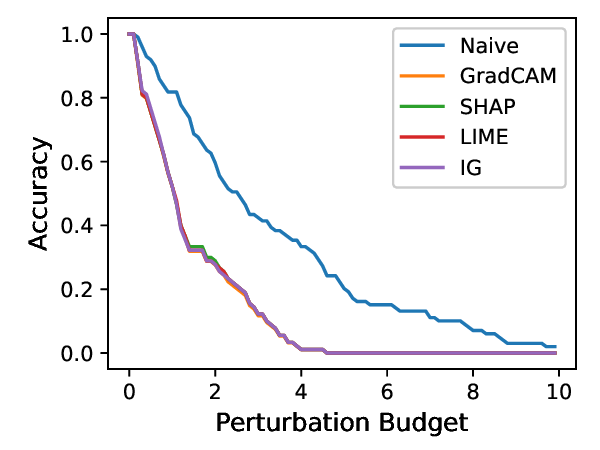}\\
    \includegraphics[width=\linewidth]{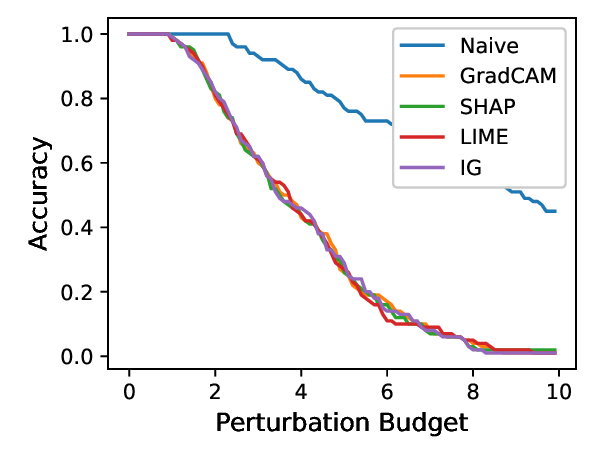}\\
    \includegraphics[width=\linewidth]{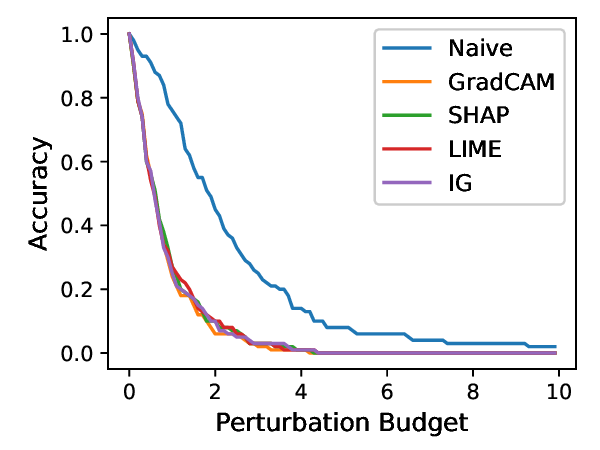}
    \caption*{(c) CUB/ResNet-50}
  \end{minipage}
  \hfill
    \begin{minipage}{0.215\textwidth}
    \includegraphics[width=\linewidth]{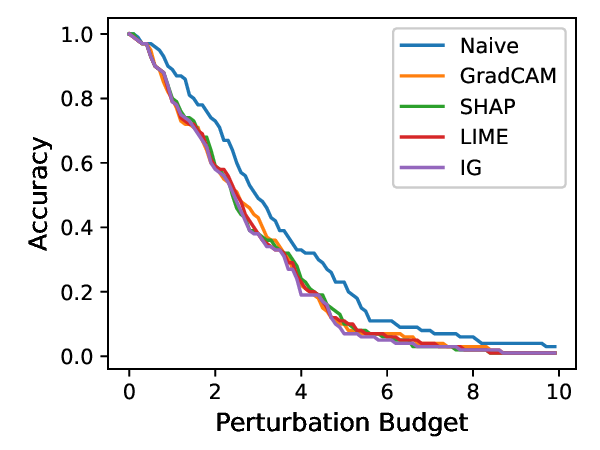}\\
    \includegraphics[width=\linewidth]{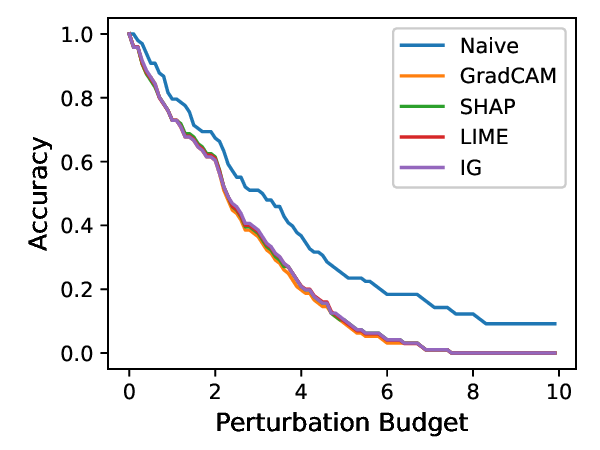}\\
    \includegraphics[width=\linewidth]{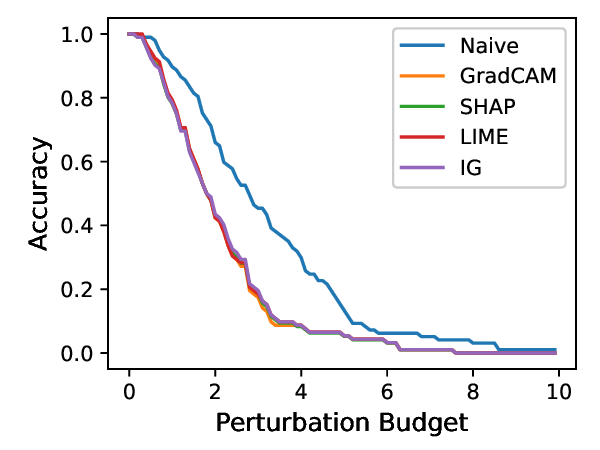}\\
    \includegraphics[width=\linewidth]{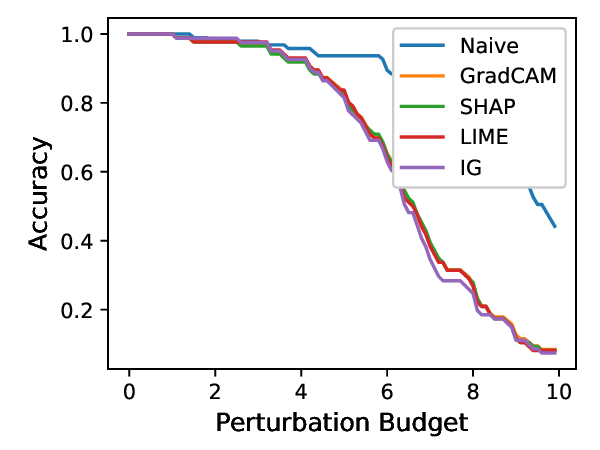}\\
    \includegraphics[width=\linewidth]{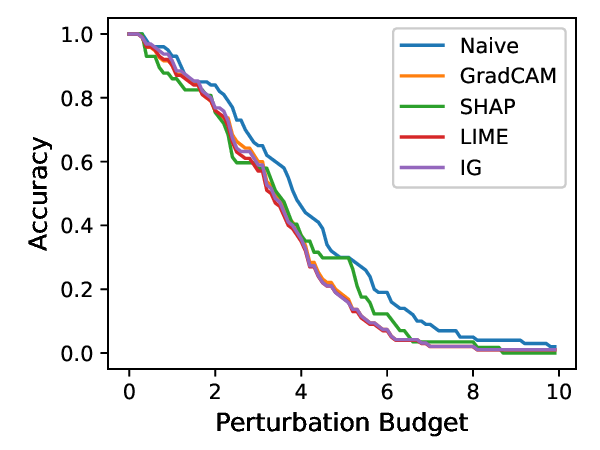}
    \caption*{(d) Stanford/ResNet-34}
  \end{minipage}
  \caption{The accuracy vs. perturbation budget curves across different datasets and explanators. The first four rows show the curves for pre-trained models, while the last row represents non-pre-trained models. The first row shows curves for the HSJA, the second row shows the curves for the Sign-Opt, the third row shows RamBoAttack, and the fourth row depicts the HSJA targeted attack. The final row shows the curves for HSJA on non-pre-trained models.}
  \label{fig:enhanced_comprehensive_perb}
\end{figure*}

\bsub{Pre-trained Setting: }
We apply our attack framework to three existing attacks: the HSJA Attack \cite{chen2020hopskipjumpattack}, the Sign-Opt Attack \cite{cheng2019sign}, and the RamBoAttack \cite{Vo2022}, implementing their naive variants as baselines for comparison. A comprehensive evaluation is conducted across various explainability methods and datasets. The results, presented in Table \ref{tab:hard_label_comprehensive}, report median $l_2$ distortions and model accuracy under a certain perturbation budget.
Our methods consistently outperform the baseline attacks. For example, in the case of HSJA on the CUB-200 dataset, applying GradCAM explanations reduced the median distortion to 0.82, compared to 2.64 for the naive attack. Figures \ref{fig:enhanced_comprehensive_query} and \ref{fig:enhanced_comprehensive_perb} depict the accuracy versus query curves and accuracy versus perturbation budget curves, respectively. Our attacks, leveraging different explanation methods, achieve lower model accuracy for the same perturbation budget while demonstrating greater efficiency by requiring fewer queries. Specifically, to reduce the robust accuracy from 1.0 to 0.66 on the CUB dataset using HSJA, the naive attack requires approximately 10,000 queries, while our low-rank adversarial attack achieves this with less than 1,000 queries—a 90\% reduction in query cost. We also evaluated the performance of targeted attacks by applying our low-rank attack framework to the HSJA targeted attack. Significant performance gains were observed; for example, on the CUB dataset, the median distortion was reduced from 9.14 to 3.64.

Additionally, our attack remains robust across different explanation methods, enabling adversaries to adopt off-the-shelf methods and thus reduce effort in practice. Interestingly, our approach is especially effective on ResNet-50 trained on the CUB dataset. This phenomenon can be explained by Figure \ref{fig:low_rank_analysis}, where ResNet-50 trained on CUB-200 exhibits a notably lower rank compared to models trained on ImageNet, offering a larger exploitable space for low-rank attacks compared to ImageNet-trained models.

\bsub{Non-pre-trained Setting: }
In the non-pre-trained setting, we extend our low-rank adversarial attack framework to HSJA attacks across various datasets and explainability methods. Table \ref{tab:hard_label_comprehensive} presents the median $l_2$ distortions and model accuracy under a perturbation budget of $\epsilon=2$. Figures \ref{fig:enhanced_comprehensive_query} and \ref{fig:enhanced_comprehensive_perb} illustrate the accuracy versus query curves and accuracy versus perturbation budget curves, respectively. Our results show that even in non-pre-trained scenarios—where the reference model architecture differs from that of the victim model, and the auxiliary dataset is completely distinct from the victim model’s training data—our attacks remain significantly more effective than baseline strategies. This demonstrates the strong transferability of the low-rank subspace across different datasets and model architectures. Notably, the ability to construct an effective low-rank subspace with minimal adversarial knowledge underscores the versatility and efficacy of our approach in boosting adversarial attack performance in different scenarios.

\subsection{Investigation on the Subspace} \label{subsec:exp_subspace}

Throughout our previous experiments, we have demonstrated that our attack remains effective even when the auxiliary data and reference model are different from the target training data and models. In this section, we provide a more in-depth ablation study that examines the key factors influencing subspace construction and discuss how these insights can guide the building of the auxiliary dataset and reference model in practice.


\begin{figure}[h]
\subfigure[Accuracy vs. Queries]{
\centering
\includegraphics[width=0.465\linewidth]{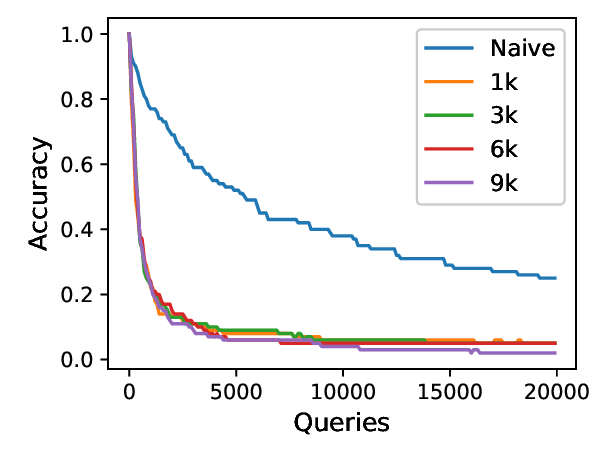}
}
\hfill
\subfigure[Accuracy vs. Perturb. Budget]{
\centering
\includegraphics[width=0.465\linewidth]{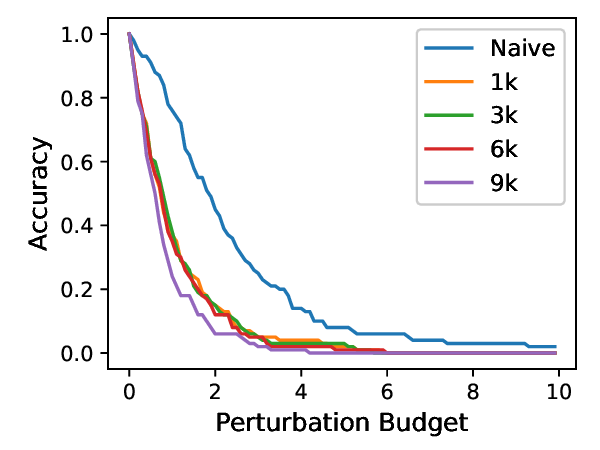}
}
\caption{Comparison of adversarial attack performance under different sizes of the auxiliary dataset. }
\label{fig:ablation_auxiliary_size}
\end{figure}

\bsub{Size of Auxiliary Dataset. }
In this section, we investigate how the size of the auxiliary datasets impacts attack efficacy. We employ the CUB-200 dataset and utilize a non-pre-trained ResNet-50 model for our experiments. To generate gradients for training the autoencoder, we vary the sizes of the selected ImageNet datasets, experimenting with sizes of 1k, 3k, 6k, and 9k images. The relationship between the auxiliary dataset size and attack performance is shown in Figure \ref{fig:ablation_auxiliary_size}. Additionally, the median $l_2$ distortions for the dataset sizes of 1k, 3k, 6k, and 9k are 0.72, 0.79, 0.72, and 0.58, respectively. Notably, the dataset containing 9k images outperforms the others in terms of attack success efficacy. Another important finding is that even with significantly constrained amount of auxiliary datasets (\textit{e.g.}, 1k images), our attack method still notably surpasses the baselines. 

\begin{figure}[h]
\subfigure[Accuracy vs. Queries]{
\centering
\includegraphics[width=0.465\linewidth]{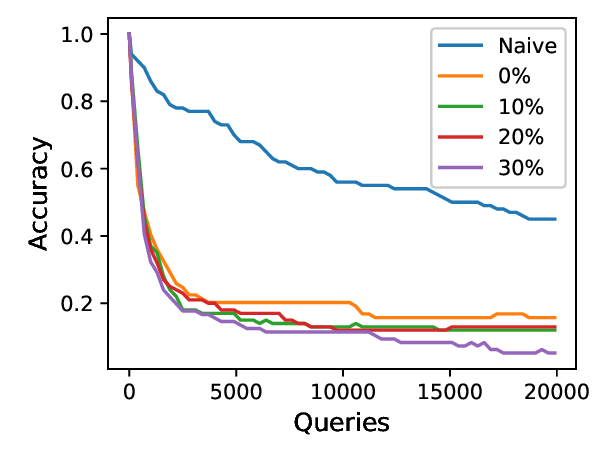}
}
\hfill
\subfigure[Accuracy vs. Perturb. Budget]{
\centering
\includegraphics[width=0.465\linewidth]{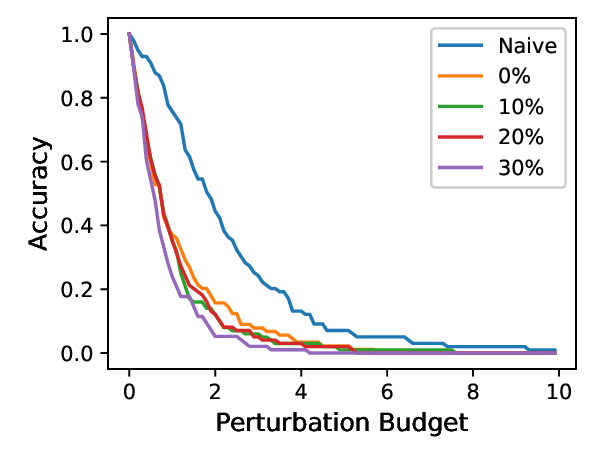}
}
\caption{Comparison of attack performance under different levels of similarity between the auxiliary and target dataset. }
\label{fig:ablation_auxiliary_similarity}
\end{figure}

\bsub{Distribution of Auxiliary Dataset. } 
Distribution is another critical factor determining the effectiveness of the constructed subspace. We hypothesize two key aspects that affect this effectiveness: (1) the distribution similarity between the auxiliary datasets and the training data of the target model, and (2) the diversity of the auxiliary datasets, as more diverse datasets can lead to learning a more general and transferable low-rank subspace. To test these hypotheses, we designed two sets of experiments. First, to evaluate the impact of distribution similarity, we use ImageNet as the auxiliary dataset and a non-pre-trained ResNet-50 trained on CUB-200 as the target model. CUB-200 specializes in fine-grained bird species classification, and ImageNet encompasses 1,000 diverse classes, including some bird-related categories (such as goldfinch, junco, and magpie). We construct auxiliary datasets by selecting varying portions (0\%, 10\%, 20\%, and 30\%) of bird-related classes from the ImageNet validation set, with the remaining samples coming from non-bird classes. The results, shown in Figure \ref{fig:ablation_auxiliary_similarity}, reveal that higher similarity improves attack effectiveness. This is expected, as similar features facilitate more accurate subspace construction. Notably, even when the overlap ratio is 0\%, our method still significantly outperforms the baseline approaches.

Second, we investigate the role of auxiliary dataset diversity. To this end, we use the following auxiliary datasets to attack CUB-200: ImageNet and Caltech-101, which are general object classification datasets; Stanford Cars, which focuses on classifying car brands; and the Large-scale CelebFaces Attributes (CelebA) dataset, which contains facial images for person recognition. The results, provided in Table \ref{tab:auxiliary_dataset}, show that general datasets perform better than specialized ones. This is likely because specialized datasets contain a narrow range of features that may not generalize well to other categories. Nonetheless, even narrow datasets perform much better than baseline methods. To summarize our explorations, if the adversary knows the target task, they can collect an auxiliary dataset from a similar domain to enhance attack effectiveness. Additionally, the adversary should use a diverse dataset, such as a multi-object classification dataset (\textit{e.g.}, ImageNet) or a combination of multiple datasets, to cover a wide range of features.

\begin{table}[t]
\small
\centering
\caption{The median $l_2$ distortion and model accuracy ($\epsilon=2$) of applying different auxiliary datasets.}
\label{tab:auxiliary_dataset}
\renewcommand{\arraystretch}{0.94}
\setlength{\belowrulesep}{1.5pt}
\setlength{\tabcolsep}{2.5pt}
\begin{tabular}{c|cc|cc}
\toprule
\hline
\multirow{2}{*}{\textbf{\begin{tabular}[c]{@{}c@{}}Auxiliary \\ Dataset\end{tabular}}} & \multicolumn{2}{c|}{\textbf{LIME}} & \multicolumn{2}{c}{\textbf{SHAP}} \\ \cline{2-5} 
                                                                                       & Median Dist.       & Accuracy      & Median Dist.      & Accuracy      \\ \hline
ImageNet                                                                               & 0.58               & 0.10          & 0.62              & 0.10          \\ \hline
Caltech-101                                                                                & 0.64               & 0.13          & 0.63              & 0.12          \\ \hline
Stanford Cars                                                                              & 0.79               & 0.20          & 0.80              & 0.21          \\ \hline
CelebA                                                                                 & 0.75               & 0.19          & 0.77              & 0.19          \\ \hline
\end{tabular}
\end{table}

\bsub{Selection of Reference Models. } 
We evaluate the impact of different reference models on attack effectiveness. Using Caltech-101 as the auxiliary dataset, we test DenseNet-121, Inception V3, and ResNet-50 as reference models to attack a Vision Transformer trained on ImageNet. The results, shown in Table \ref{tab:reference_model}, indicate that our method remains effective across different reference models. In practice, adversaries could try multiple off-the-shelf reference models to optimize attack efficacy and efficiency.

\begin{table}[t]
\small
\centering
\caption{The median $l_2$ distortion and model accuracy ($\epsilon=2$) of applying different reference models.}
\label{tab:reference_model}
\renewcommand{\arraystretch}{0.94}
\setlength{\belowrulesep}{1.5pt}
\setlength{\tabcolsep}{2.5pt}
\begin{tabular}{c|cc|cc}
\toprule
\hline
\multirow{2}{*}{\textbf{\begin{tabular}[c]{@{}c@{}}Reference\\ Model\end{tabular}}} & \multicolumn{2}{c|}{\textbf{LIME}} & \multicolumn{2}{c}{\textbf{SHAP}} \\ \cline{2-5} 
                                                                                    & Median Dist.       & Accuracy      & Median Dist.      & Accuracy      \\ \hline
DenseNet-121                                                                            & 1.32               & 0.28          & 1.35              & 0.30          \\ \hline
Inception V3                                                                        & 1.43               & 0.30          & 1.43              & 0.31          \\ \hline
ResNet-50                                                                           & 1.24               & 0.24          & 1.26              & 0.25          \\ \hline
\end{tabular}
\end{table}

\subsection{Ablation Study}

\begin{figure}[h]
\subfigure[Accuracy vs. Queries]{
\centering
\includegraphics[width=0.465\linewidth]{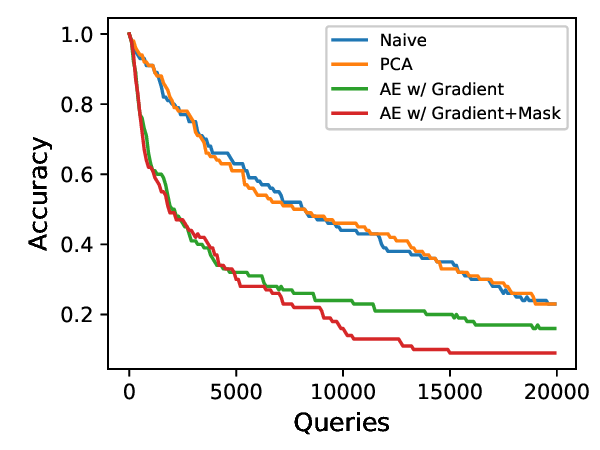}
}
\hfill
\subfigure[Accuracy vs. Perturb. Budget]{
\centering
\includegraphics[width=0.465\linewidth]{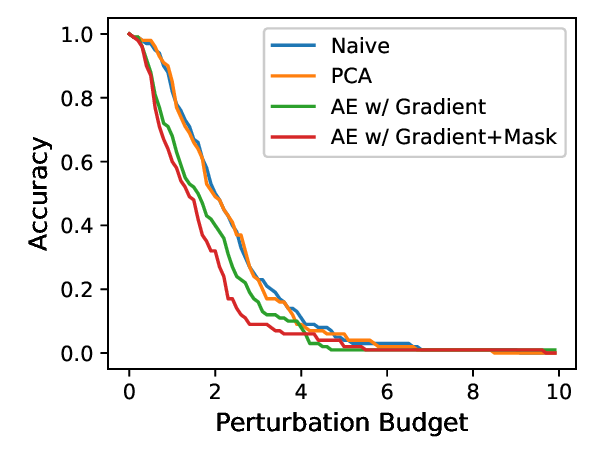}
}
\caption{Comparison of adversarial attack performance between different subspace optimization methods.}
\label{fig:ablation_subspace}
\end{figure}

\bsub{Effects of Diverse Attack Components on Subspace Optimization. }
In this part, we investigate the impacts of different components in low-rank subspace construction. Specifically, to evaluate the effectiveness of AE, we introduce a baseline using the PCA algorithm to learn the subspace. To further assess the effectiveness of explanations, we compare the AE using only gradients (AE w/ Gradient) and the AE using both gradients and explanation masks (AE w/ Gradient + Mask) for training. For the PCA baseline, we utilize the same 9,000 images and DenseNet-121 to generate gradient vectors as used in the AE. We employ the scalable approximate PCA algorithm \cite{halko2011finding} to extract the top 9,024 major components, defining the intrinsic subspace. Figure \ref{fig:ablation_subspace} shows the comparison of these methods. The results reveal that PCA exhibits attack performance similar to the naive baseline attack, yet it significantly underperforms AE, demonstrating the efficacy of the AE in capturing the non-linear gradient space. Moreover, the integration of gradients and explanations outperforms the gradient-only approach in both efficacy and efficiency, highlighting the advantages of explanations in capturing essential gradient information.


\begin{figure}[h]
\subfigure[Accuracy vs. Queries]{
\centering
\includegraphics[width=0.465\linewidth]{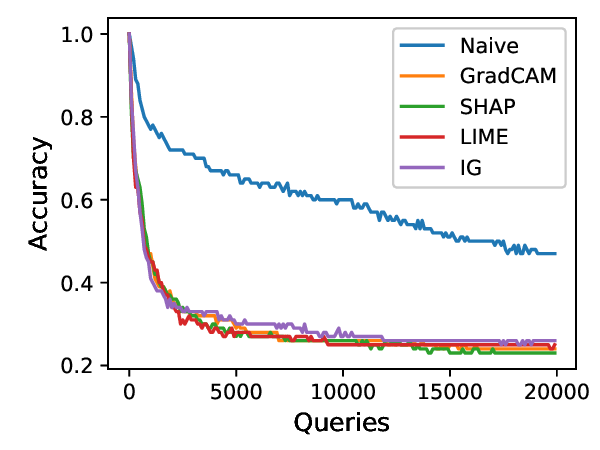}
}
\hfill
\subfigure[Accuracy vs. Perturb. Budget]{
\centering
\includegraphics[width=0.465\linewidth]{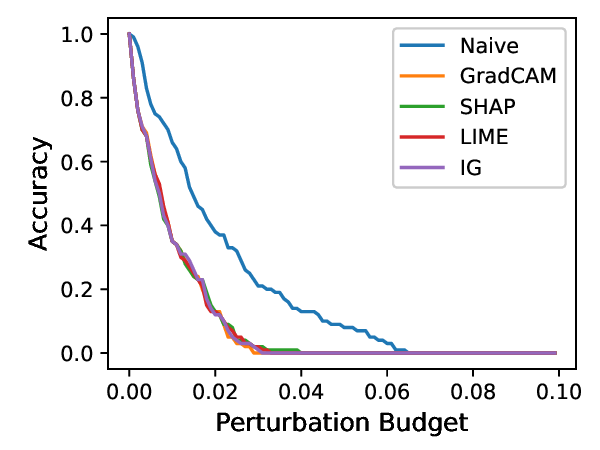}
}
\caption{Adversarial attack performance under $l_\infty$ norm. The accuracy vs. queries curve is obtained using $\epsilon=4/255$.}
\label{fig:eval_linf}
\end{figure}

\begin{figure*}[h]
    \centering
    \includegraphics[scale=0.37]{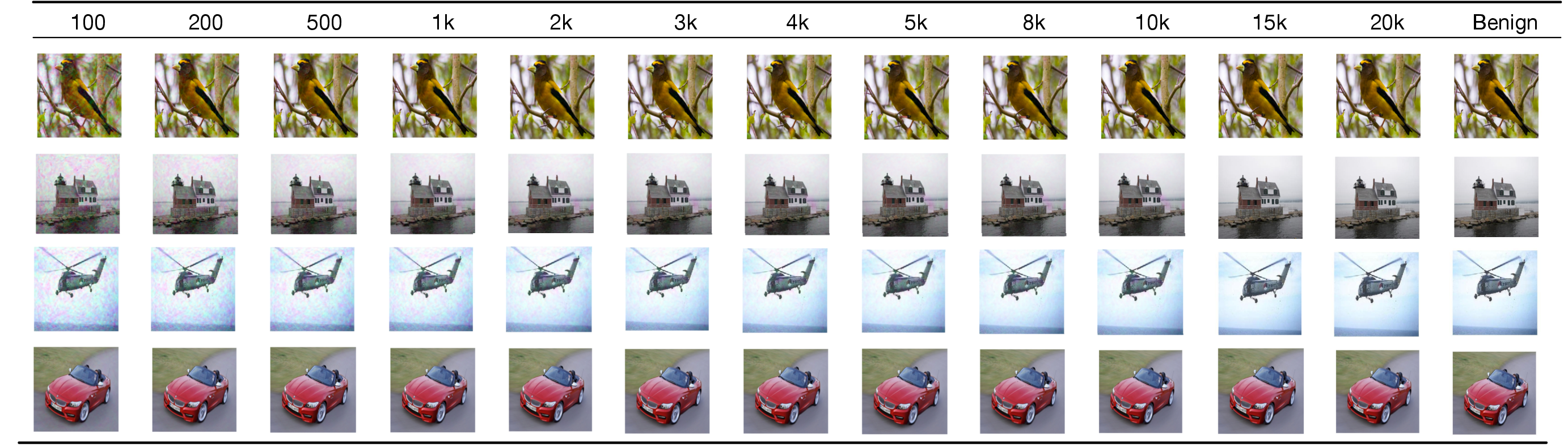}
    \caption{Visualized trajectories of our proposed attacks on randomly selected images in CUB-200, ImageNet, Caltech-101, and Standford Car. 1st-12th columns: generated adversarial candidates when query number is 100, 200, 500, 1K, 2K, 3K, 4K, 5K, 8K, 10K, 15K, and 20K. 13th column: The benign image. 1st row: low-rank attacks on CUB-200. 2nd row:  low-rank attacks on ImageNet. 3rd row: low-rank attacks on Caltech-101. 4th row: low-rank attacks on Standford Car.}
\label{fig:demo}
\end{figure*}

\bsub{Evaluation of $l_\infty$ norm: }
We additionally conduct evaluations under the $l_\infty$ norm, applying our subspace optimization to HSJA attacks and performing experiments on the CUB-200 datasets. The outcomes, presented in Figure \ref{fig:eval_linf}, demonstrate the efficacy and efficiency of our method in scenarios under the $l_\infty$ norm. To demonstrate that the perturbations are imperceptible, we present the visualized trajectories of our proposed attacks in Figure \ref{fig:demo}.

\begin{figure}[h]
\subfigure[TRADES]{
\centering
\includegraphics[width=0.465\linewidth]{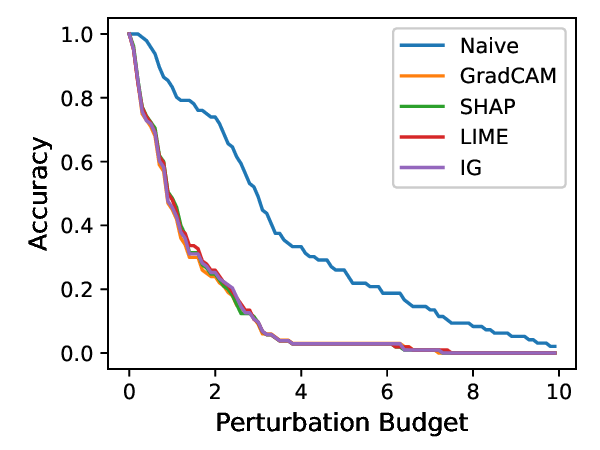}
}
\hfill
\subfigure[Distillation]{
\centering
\includegraphics[width=0.465\linewidth]{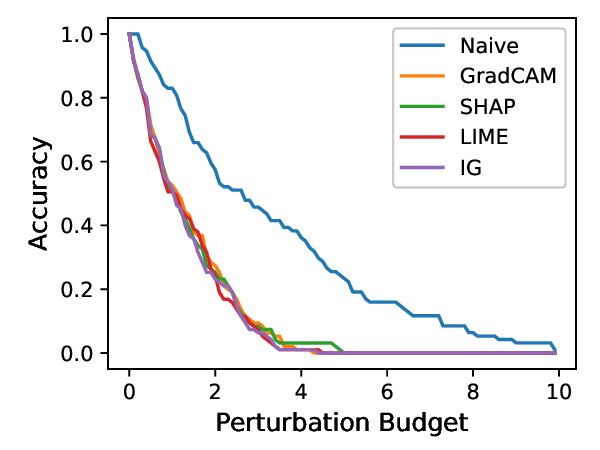}
}

\subfigure[Salman et al.]{
\centering
\includegraphics[width=0.465\linewidth]{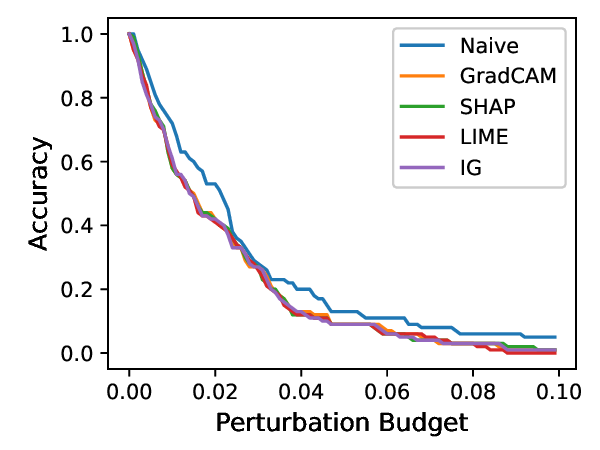}
}
\hfill
\subfigure[Chen et al.]{
\centering
\includegraphics[width=0.465\linewidth]{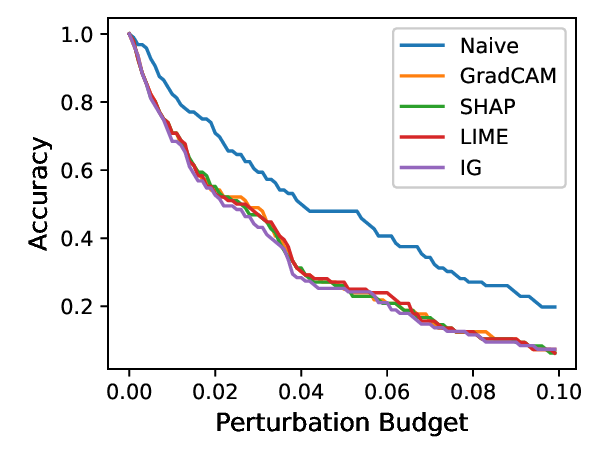}
}
\caption{The accuracy vs. perturbation budget curves of the four adversarially trained robust models.}
\label{fig:robust_eval}
\end{figure}

\begin{table*}[h]
\small
\centering
\caption{The median distortion and model accuracy under adversarial attacks on four adversarially trained robust models. The accuracy under the $l_2$ norm is evaluated with $\epsilon=2$, while the accuracy under the $l_\infty$ norm is evaluated with $\epsilon=4/255$.}
\label{tab:robust_eval}
\renewcommand{\arraystretch}{0.94}
\setlength{\belowrulesep}{1.5pt}
\setlength{\tabcolsep}{3pt}
\begin{tabular}{c|c|cc|cc|cc|cc|cc}
\toprule
\hline
\multirow{2}{*}{\textbf{Distance}} & \multirow{2}{*}{\textbf{Robust Method}} & \multicolumn{2}{c|}{\textbf{Naive HSJA}} & \multicolumn{2}{c|}{\textbf{GradCAM}} & \multicolumn{2}{c|}{\textbf{IG}} & \multicolumn{2}{c|}{\textbf{LIME}} & \multicolumn{2}{c}{\textbf{SHAP}} \\ \cline{3-12} 
                          &                                         & Median Dist.     & Acc.     & Median Dist.    & Acc.   & Median Dist.    & Acc.   & Median Dist.  & Acc.  & Median Dist.  & Acc. \\ \hline
\multirow{2}{*}{$l_2$}    & TRADES \cite{zhang2019theoretically}                                 & 2.95             & 0.74         & 0.88            & 0.24       & 0.90            & 0.25       & 0.91          & 0.26      & 0.92          & 0.25     \\
                          & Distillation  \cite{papernot2016distillation}                          & 2.65             & 0.57        & 1.16           & 0.28       & 1.05           & 0.23       & 1.05          & 0.24     & 1.02          & 0.25     \\ \hline
\multirow{2}{*}{$l_\infty$} & Salman et al. \cite{salman2020adversarially}                              & 0.02             & 0.58         & 0.01            & 0.47       & 0.01            & 0.46       & 0.01          & 0.46      & 0.01          & 0.47     \\
                          & Chen et al. \cite{chen2024data}                             & 0.04             & 0.76         & 0.03            & 0.59       & 0.02            & 0.57       & 0.03          & 0.58      & 0.03          & 0.60     \\ \hline
\end{tabular}
\end{table*}

\subsection{Evaluation on Robust Models} \label{subsec:robust_model}

In this section, we investigate the effectiveness of our proposed low-rank adversarial attacks against adversarially robust models. To this end, we have selected four representative defense mechanisms against adversarial examples. These include the TRADES  \cite{zhang2019theoretically} and the defensive distillation \cite{papernot2016distillation} as mechanisms evaluated under the $l_2$ norm, alongside approaches by Salman et al. \cite{salman2020adversarially} and Chen et al. \cite{chen2024data}, which operate under the $l_\infty$ norm. 

TRADES \cite{zhang2019theoretically} proposes a new formulation of adversarial training by optimizing a regularized surrogate loss to achieve an optimal trade-off between robustness and accuracy. Since the pre-trained weights are not released, we follow the codes and parameters in the official repository to perform adversarial training on ResNet50 models on CUB-200 datasets with $\epsilon=5$. Defensive Distillation \cite{papernot2016distillation} employs soft probability vectors from the original network to train a robust model with an identical architecture. We use a pre-trained ResNet-50 on the CUB-200 dataset, setting the distillation temperature $T = 2$.
Salman et al. \cite{salman2020adversarially} systematically investigate the adversarial robustness of ImageNet models. we select the released ResNet18 models trained under an $l_\infty$ norm setting with $\epsilon=4/255$. Chen et al. \cite{chen2024data} propose a novel adversarial training scheme to achieve optimal robust accuracy while notably reducing the computational cost of training. We use the released Wide Residual Net with a width of 10 and a depth of 34 \cite{zagoruyko2016wide} trained on ImageNet with $\epsilon=8/255$ as our target model.
We apply the low-rank attack framework to the HSJA attacks with various explanation methods to evaluate the attack effectiveness. The evaluation results are given in Figure \ref{fig:robust_eval} and Table \ref{tab:robust_eval}, demonstrating the effectiveness of our low-rank attacks against adversarially trained models. While TRADES provides stronger protection compared to defensive distillation since the latter has been shown to be vulnerable to attacks \cite{carlini2016defensive}, our methods remain effective even against these robust models.

\section{Related Work} \label{sec:related_work}

\bsub{Adversarial Attacks and Defenses. }
Since the first introduction of adversarial examples \cite{szegedy2013intriguing}, various attacks have been proposed. Goodfellow et al. introduced the Fast Gradient Sign Method (FGSM) \cite{goodfellow2014explaining}, which generates adversarial examples through a one-step optimization process on the input image towards the gradient ascent path. Following this work, various white-box attacks are proposed, including the Basic Iterative Method (BIM) \cite{kurakin2018adversarial}, Projected Gradient Descent (PGD) \cite{madry2017towards}, and the C\&W attack \cite{carlini2017towards}. 
Subsequent research expanded these attacks into more practical threat models, including score-based black-box settings with methods like ZOO \cite{chen2017zoo}, SPSA \cite{uesato2018adversarial}, and Bandits \cite{ilyas2018prior}, and hard-label settings with methods like Boundary Attack \cite{brendel2018decision}, Opt \cite{cheng2018query}, Sign-Opt \cite{cheng2019sign}, HSJA \cite{chen2020hopskipjumpattack}, and RamBoAttack \cite{Vo2022}. Concurrently, the defenses have been extensively explored, including adversarial training \cite{li2019improving,pang2019improving,salman2020adversarially,zhang2019theoretically}, defensive distillation \cite{papernot2016distillation}, input transformation \cite{samangouei2018defense}, etc. However, adaptive attacks have been shown to successfully break many of these defense mechanisms \cite{athalye2018obfuscated,carlini2019evaluating,tramer2020adaptive}, highlighting the ongoing challenge of developing robust defenses against adversarial attacks.

\bsub{Low Rank Adaptation (LoRA). } 
In recent years, the parameter scales of pre-trained language models have increased dramatically, following the success of scaling laws in language models \cite{kaplan2020scaling}. Despite this growth, large language models (LLMs) still face limitations in performing certain downstream tasks due to the knowledge boundary, which necessitates task-specific fine-tuning \cite{mao2024survey}. However, fine-tuning all the parameters of an LLM is computationally prohibitive. To address this challenge, various parameter-efficient fine-tuning (PEFT) methods have been proposed \cite{ding2023parameter}. Among these, LoRA \cite{hulora2022} stands out for its ability to introduce no additional inference latency. LoRA builds on the insight that over-parameterized models reside on a low intrinsic dimensional space. It achieves efficiency by updating only the dense layers using low-rank matrices, significantly reducing the number of trainable parameters. This approach is highly memory and computation efficient, as it updates only a small subset of the model's parameters. For instance, Pan et al. \cite{pan2024lisa} compare full fine-tuning to LoRA fine-tuning on the LLaMA2-7B model \cite{touvron2023llama}. They report that full fine-tuning requires approximately 60GB of memory, which exceeds the capacity of most consumer GPUs. In contrast, LoRA fine-tuning only requires around 23GB, fitting comfortably within the memory limits of most consumer GPUs. LoRA has been widely adopted in various downstream tasks to significantly reduce computational resource requirements while maintaining performance, including text classification \cite{li2023label}, automated program repair \cite{silva2023repairllama}, multi-language to Python code translation \cite{pan2023stelocoder}, and federated learning \cite{cho2023heterogeneous,babakniya2023slora}.

\section{Conclusion} \label{sec:conclusion}


In this paper, we conducted a systematic rank analysis, both theoretical and empirical, of adversarial perturbations, demonstrating that adversarial perturbations are intrinsically low-rank. Leveraging this property, we crafted more efficient black-box adversarial examples. Our approach first utilizes a reference model and auxiliary data to guide the projection of gradients into a low-dimensional subspace. By confining the perturbation search to this low-rank subspace, we significantly improved the efficiency and effectiveness of black-box adversarial attacks. Extensive evaluation across a variety of attack methods, benchmark models, and datasets, demonstrated substantial performance improvements over existing baseline attacks.





\newpage
\newpage

\bibliographystyle{ieeetr} 

\bibliography{reference}{}

\appendices


\section{Experimental Settings} \label{appendix:exp_setting}

\bsub{Datasets. } We use the following datasets during the experiment: 
\noindent \textit{ImageNet \cite{deng2009imagenet}:} The ImageNet dataset is one of the most widely used datasets for image classification benchmarks. It spans 1000 object classes and contains 1,281,167 training images, 50,000 validation images, and 100,000 test images.

\noindent \textit{CUB-200 \cite{wah2011caltech}}: The Caltech-UCSD Birds-200-2011 dataset contains 11,788 images of 200 subcategories belonging to birds, 5,994 for training and 5,794 for testing.

\noindent \textit{Stanford Cars \cite{krause20133d}}: The Stanford Cars dataset contains 16,185 images of 196 classes of cars. The data is split into 8,144 training images and 8,041 testing images, where each class has been split roughly in a 50-50 split.

\noindent \textit{Caltech-101 \cite{fei2004learning}}: The Caltech-101 dataset contains a total of 9,146 images, split between 101 distinct object categories. For each object category, there are about 40 to 800 images, while most classes have about 50 images.

\noindent \textit{CelebA \cite{liu2015faceattributes}}: The Large-scale CelebFaces Attributes (CelebA) dataset is a large-scale face dataset containing 202,599 images of 10,177 identities. The dataset is typically split into 162,770 images for training, 19,867 for validation, and 19,962 for testing, ensuring coverage of a wide variety of facial appearances and attributes.

\bsub{Adversarial Attack Settings. } 
To implement the C\&W attack, we follow the popular Torchattacks library \cite{kim2020torchattacks}. We use the default parameters with 1000 iterations and a learning rate of 0.01. 
For the Bandits Attack, we employ its variant with time and data-dependent priors. We utilize its official GitHub repository \cite{ilyas2018prior} and run the attacks with a maximum of 10,000 queries using default parameters. We resolved a bug in the official repository related to computing correct labels on unnormalized images (issue \# 3).
In the case of the HSJA attack, we adhere to the methodology outlined in the widely-recognized Foolbox library \cite{rauber2017foolbox}. Specifically, we set the number of generated adversarial examples for each iteration to 50, with the initial iteration generating 100 adversarial examples, and we set $\gamma=0.01$. For the Sign-Opt attack, our implementation is based on the Attackbox library \cite{cheng2018query}. We configure the parameters as follows: $k=200$, $\alpha=0.1$, and $\beta=1$. Lastly, to implement the RamBoAttack, we employ its official implementation \cite{Vo2022} and run it with a maximum of 20,000 queries.

\bsub{Victim Model Training. } For all models, the input images are resized to dimensions (224, 224, 3). Specifically, we conduct training over 200 epochs, utilizing a learning rate of 0.001, momentum set to 0.9, and applying weight decay at a rate of 0.0005.

\section{Comparison with Baseline Attack} \label{appendix:comp_baselines}

In this section, we quantitatively compare our proposed attack with that of Esmaeili et al. \cite{esmaeili2023low}. We use EfficientNet as the target model and randomly select the same 100 ImageNet images for both attacks. We then apply our low-rank framework to the HSJA attack, following the same experimental and evaluation protocols described in Section \ref{subsec:exp_setup}. The results, presented in Table \ref{tab:lsdat_attack}, show that our method achieves substantially lower $\ell_2$ distortions and also significantly reduces the robust accuracy compared to Esmaeili et al. At a perturbation budget of $\epsilon=2$, Esmaeili et al.’s approach barely decreases the model’s accuracy, whereas our attack dramatically lowers it. While their method may perform adequately with very low query budgets, it fails to converge to optimal solutions even when given additional queries. This convergence to suboptimal solutions results in highly perceptible adversarial examples with notably lower quality. 

\begin{table}[t]
\small
\centering
\caption{The median $l_2$ distortion and model accuracy under $\epsilon=2$ of our attack and attack proposed by Esmaeili et al.}
\label{tab:lsdat_attack}
\renewcommand{\arraystretch}{0.99}
\setlength{\belowrulesep}{2pt}
\setlength{\tabcolsep}{6pt}
\begin{tabular}{c|c|c}
\toprule
\hline
\textbf{Method}                         & \textbf{Median Distance} & \textbf{Accuracy ($\epsilon=2$)} \\ \hline
Esmaeili et al. \cite{esmaeili2023low} & 16.33           & 0.97                    \\ \hline
Ours w/ GradCAM                & 1.40            & 0.31                    \\ \hline
Ours w/ IG                     & 1.30            & 0.26                    \\ \hline
Ours w/ LIME                   & 1.49            & 0.27                    \\ \hline
Ours w/ SHAP                   & 1.39            & 0.32                    \\ \hline
\end{tabular}
\end{table}

\section{Architecture of the Autoencoder} \label{appendix:autoencoder}

The architecture of the autoencoder is designed to compress and subsequently reconstruct input data through a series of encoding and decoding operations. Specifically, it is composed of four encoder blocks and four decoder blocks. Each encoder block employs two 3x3 convolutional layers, with each convolutional operation followed by a Rectified Linear Unit (ReLU) activation. Subsequent to these convolutions, a 2x2 max-pooling layer is applied, effectively halving the spatial dimensions of the feature maps, thereby reducing the data representation size. On the decoding side, each decoder block initiates with a 2x2 transposed convolution, serving to upscale the feature maps. This is followed by two 3x3 convolutional layers, each accompanied by a ReLU activation function. The final stage of the decoder involves a 1x1 convolution, which is employed to generate the final reconstructed output, mapping the processed features back to the original data space. This structure allows the AE to learn an efficient representation of the input data, facilitating data compression.

\end{document}